\documentclass[11pt]{article}

\usepackage{amsmath,multirow}    
\usepackage[table]{xcolor}
\usepackage[T1]{fontenc}
\usepackage[utf8]{inputenc}

\usepackage{epstopdf}
\usepackage{amsmath}
\usepackage{epsf}
\usepackage{graphicx}

\usepackage{amsfonts}
\usepackage{amsthm}
\usepackage{amsmath}
\usepackage{mathrsfs}
\usepackage{amssymb, stmaryrd}
\usepackage{graphicx,algorithm, algorithmic}
\usepackage{caption}
\RequirePackage[colorlinks,linkcolor=blue,citecolor=blue,urlcolor=blue]{hyperref}
\usepackage{latexsym,epsf,dsfont,color,eurosym,mathrsfs, url}
\mathchardef\mhyphen="2D
\usepackage{graphicx}

\usepackage[round,colon,authoryear]{natbib}

\DeclareMathOperator*{\argmin}{\arg\!\min}
\DeclareMathOperator*{\argmax}{\arg\!\max}

\usepackage{bbm}

\newtheorem{remark}{Remark}
\usepackage{mathrsfs,bbm}

\usepackage{tikzsymbols,tikz,pgfplots}
\usetikzlibrary{snakes}

\usepackage{graphicx}
\usepackage{caption,authblk}
\usepackage{subcaption}

\theoremstyle{plain}
\newtheorem{define}{Definition}
\theoremstyle{plain}
\newtheorem{thm}{Theorem}
\theoremstyle{plain}
\newtheorem{lemma}{Lemma}
\theoremstyle{plain}
\theoremstyle{plain}
\newcommand{\tab}{\hspace*{2em}}

\renewcommand{\P}{\mathbb{P}}
\newcommand{\Var}{\textsf{\text{Var}}}

\renewcommand{\P}{\mathbb{P}}
\newcommand{\R}{{\mathbb{R}}}
\newcommand{\E}{\mathbb{E}}

\title {  Nonparametric Modeling of Higher-Order Interactions via Hypergraphons}
\author{Krishnakumar Balasubramanian}
\affil{\texttt{kbala@ucdavis.edu}}
\affil{Department of Statistics, University of California, Davis.}
\date{}
\begin{document}
\maketitle
\begin{abstract}
We study statistical and algorithmic aspects of using hypergraphons, that are limits of large hypergraphs, for modeling higher-order interactions. Although hypergraphons are extremely powerful from a modeling perspective, we consider a restricted class of Simple Lipschitz Hypergraphons (SLH), that are amenable to practically efficient estimation. We also provide rates of convergence for our estimator that are optimal for the class of SLH. Simulation results are provided to corroborate the theory. 
\end{abstract}

\section{Introduction}\label{sec:intro}
Let $V= \{1, \ldots, n \}$ be a set of $n$ items that could represent for example, people in a social network, genes in a biological network or researchers in academic networks. Models of interaction among the $n$ items could be conveniently represented in the form of a graph or a hypergraph, $G(V,E)$, where the items form the nodes of the graph and the hyperedge set $E$ represents the interactions among the items. Network datasets that capture such complex interactions between a set of objects are becoming increasingly prevalent in several scientific fields. Developing realistic generative models for such networks is a challenging problem that has been an active subject of research across diverse fields spanning from statistics, physics, computer science; see~\cite{kolaczyk2009statistical, goldenberg2010survey, battiston2020networks} for comprehensive overview.   

A majority of the existing work has focussed on the case of modeling \emph{pairwise} interactions.  In this situation, the edge set models interactions between two nodes at a time, by means of presence or absence of a link. In several modern applications, such pairwise interactions do not completely characterize the complex interactions existing among the items. Often times it is more meaningful to consider higher-order interactions~\citep{bonacich2004hyper, benson2016higher}.  For example,~\cite{agarwal2005beyond} provided convincing empirical evidence showing that going beyond pairwise interactions helps in computer vision applications; a hypergraph based approach for graph matching was provided in~\cite{duchenne2011tensor}. Yet another application is the study of complex networks, where modeling intricate interdependencies between multiple networks are typically represented via hypergraphs~\citep{ghoshal2009random,zlatic2009hypergraph, michoel2012alignment, kivela2014multilayer} so as to capture the higher-order interactions amongst the different networks. \textcolor{black}{We refer the interested reader to~\cite{battiston2020networks}, for a detailed survey of several existing models of higher-order interactions and their applications to various scientific fields.}

As a simple yet concrete example for the limitations of the just using pairwise interactions, consider the following co-author citation toy network. Consider a simple set of five authors $\{ A, B, C, D, E\}$ and assume that the set of co-author relationship among the authors is as follows: $(A,B), (B,E), (A,E) (C,D,E)$. That is the five authors wrote five papers in total with the above set of authors for each paper. If we represent this network via a graph, with edges representing if two authors have collaborated or not, we  get a graph with edge set $\{ (A,B), (B,E), (A,E), (C,E), (C,D), (D,E) \}$. We now see that the crucial coauthor information is lost by such representation - one could wrongly interpret that authors $A,B,E$ have co-authored a paper together or one could fail to conclude that authors $C,D,E$ co-authored a paper together. This highlights the limitation of modeling such co-author data set via simple graphs. Clearly if we were to model this data as a hypergraph, we do not run into such issues. A more comprehensive co-author citation network was recently considered in~\cite{ji2016} based on papers published in the Statistics community. The need for modeling higher-order interaction for that dataset was in particular also suggested in the discussion by ~\cite{karwa2016}, following the publication of~\cite{ji2016}. 

\textcolor{black}{Motivated by such problem with pairwise interaction networks or graph-based networks, recently~\cite{ghoshdastidar2015consistency} considered a model for generating hypergraphs. Specifically, they considered a generalization of a stochastic block models (we refer the reader to~\cite{abbe2017community} for a detailed overview) developed for the case of graphs to the hypergraph setting. Furthermore~\cite{florescu2016spectral} considered a special case of hypergraphs with a bipartite structure and suggested interesting algorithmic conjectures. Algorithmic results were also provided in~\cite{ghoshdastidar2016uniform,ahn2018hypergraph,ke2019community}, for community detection in hypergraphs based on tensor methods. While the above works invariably work in the realm of stochastic block modeling, there has been other approaches to model hypergraph data as well. A $\beta$-model and a Latent Class Analysis (LCA) based model (focussing on clustering) was proposed in~\cite{stasi2014beta} and~\cite{ng2018model} respectively. Furthermore,~\cite{turnbull2019latent} recently proposed a latent space model for random hypergraphs based on concepts from computational topology. \textcolor{black}{We will revisit the above models in Section~\ref{sec:slh} for a brief discussion on their connection to the approach that we propose in this work.} Furthermore, a random geometric model for random hypergraph was also proposed in~\cite{lunagomez2017geometric} focussing on graphical modeling. A significant drawback of the above approaches for modeling higher-order interactions is that they are predominantly parametric models. It is well known that nonparametric models offer increased modeling flexibility at the expense of requiring larger sample sizes in standard regression and density estimation problems. Thus providing such nonparametric models for networks is extremely appealing, particularly in the context of modeling large graphs, as the nodes in the graphs corresponds to samples in the context of traditional regression models.}

\textcolor{black}{In this work, we leverage the theory of large hypergraph limits and propose to use hypergraphons~\citep{gowers2007hypergraph, elek2012measure, lovasz2012large, zhao2015hypergraph}, as a nonparametric model to capture $m$-uniform higher-order interactions in networks. Modeling such $m$-uniform interactions arises, for example, in the context of protein network alignment problem used to represent interactions across different organisms. In this context, we are given two graphs $G_1$ and $G_2$ whose vertices are connected by a bipartite graph $B$. Based on this, a 4-uniform alignment hypergraph is formed with an hyperedge connecting the nodes $i,j,k$ and $l$ if and only if the notes $i$ and $j$ are connected in the graph $G_1$ and nodes $k$ and $l$ are connected in the graph $G_2$. We refer to the interested to reader to~\cite{michoel2012alignment} for more details.} As we will see in the rest of the paper, in the limit of large nodes, the difference between the expressibility of parametric (for example, block hypergraph models) and nonparametric models (the proposed hypergraphons) for modeling $m$-uniform higher-order interactions is significant. This is different from the case of modeling pairwise interactions via stochastic block models and graphons for graph-valued networks. Roughly speaking while graphs are represented by adjacency matrices (second-order tensors) and the corresponding graphons are two-dimensional functions, this is not true for hypergraphs and hypergraphons. An $m$-uniform hypergraph could be represented by $m$-th order tensor, whereas the corresponding hypergraphon is represented by a $2^m-2$ dimensional function. This expressive power comes at an increased statistical and computational price.

In order to facilitate efficient estimation and computation, we restrict ourself to a class of Simple Lipschitz Hypergraphons (SLH). Furthermore, Lemma 10 and Theorem 2 in~\cite{kallenberg1999multivariate} provide guarantees for approximating general hypergraphons with simple hypergraphons, which provides further motivation for understanding this class of hypergraphons statistically and computationally. For this class of hypergraphons, we propose an estimator along with its rates of convergence. \textcolor{black}{The proposed estimator is based on approximating this class of hypergraphons with a parametric stochastic hypergraph block models with appropriately selected number of blocks. Indeed such an approach yields rate-optimal estimators for the case of graphons~\citep{gao2015rate, klopp2015oracle}.} Unfortunately, from a computational perspective the estimator is non-convex and hence NP-hard to compute in the worst case. We provide an algorithm which is based on a well-motivated heuristic, and show thorough simulations that it works well in practice. \textcolor{black}{We also mention here that the approximation result provided in~\cite{kallenberg1999multivariate} is existential, not entirely constructive and provided in a weaker metric. It is an extremely interesting open problem to provide a constructive proof of such a result in a stronger metric, so that one could find the optimal simple hypergraphon approximation for a given hypergraphon. Such a result would constructively quantify the efficiency of approximation from a practical perspective.}

\textcolor{black}{Finally, the modern theory of large hypergraph limits has been established through the analytic regularity approach, for example, in~\cite{gowers2007hypergraph, zhao2015hypergraph}. However, a probabilistic approach based on node-exchangeability has also been examined, for example, in~\cite{hoover1979relations, aldous1981representations, kallenberg1999multivariate} to study limits of large hypergraphs. Recently, a closely related concept of edge-exchangeability has been proposed and leveraged by~\cite{crane2018edge, campbell2018exchangeable, dempsey2019hierarchical} to propose models of higher-order interactions. The above works do not rigorously analyze the statistical estimation procedure associated with the model. It is interesting to explore further relationships and dissimilarities between the proposed hypergraphon model and edge-exchangeable models as future work. Furthermore, it is extremely interesting to quantify the degree of expressibility offered by the two approaches, and compare the statistical and computational complexity of estimation in the different models.}


\subsection{Notations}\label{sec:notation}
We denote by $[n] = \{ 1, \ldots, n \}$, the set of integers from $1$ to $n$ and $ {n\choose k}$ to denote the number of ways of selecting $k$ objects out of $n$. Furthermore, we use $\mathfrak{S}_n$ to be the set of all permutations of the set $[n]$. We denote a vector $v$ in $d$-dimensional Euclidean space by small case letters $a \in \mathbb{R}^d$. Similarly we denote matrices by upper case letters $A \in \mathbb{R}^{n \times n}$. A bold upper case letter $\mathbf{A} \in \mathbb{R}_m^{n \times n \times \ldots \times n}$ corresponds to an $m^{th}$ order tensor. The $(j_1, \ldots, j_m)-th$ entry of the tensor is denoted as $\mathbf{A}_{j_1,j_2, \ldots, j_m}$. For a tensor $\mathbf{A}$ we use $\|\mathbf{A} \|_F$ and $\|\mathbf{A} \|_\infty$ to denote the standard Frobenius norm and the max-norm (maximum value of its entries) respectively.

Furthermore, whenever there is no confusion, to avoid notation overload (mostly in the proofs) we also use the following notation to index the entries of the tensor. Let $j = (j_1, \ldots, j_m)$ with each $j_i \in [n]$ so that $j \in [n]^m$. Whenever it indexes a tensor $\mathbf{A}$ as $\mathbf{A}_j$, it denotes the $(j_1, \ldots, j_m)-th$ entry of the tensor. Hence we have $\mathbf{A}_j=\mathbf{A}_{j_1, \ldots, j_m}$. Furthermore, we use $\sigma(j)$ to denote the set of all permutations of a given $j \in [n]^m$. Here $\sigma \in \mathfrak{S}_m$.  We also denote multiple summations like $\overset{n}{\underset{j_1=1}{\sum}} \cdots \overset{n}{\underset{j_m=1}{\sum}}$ as $\underset{j \in [n]^m}{\sum}$. For the sake of brevity, in the rest of the paper $z(j)$ represents $\left(z(j_1), z(j_2), \ldots, z(j_m)\right)$, when $j \in [n]^m$, and similarly for $z^{-1}(a)$ when $a \in [k]^m$. We also need the following definition of collapsing a tensor to a matrix.
\begin{define}[\textsc{Tensorcollapse}]
For a tensor $\mathbf{A} \in \mathbb{R}_m^{n \times n \times\ldots n} $, we define the operation of collapsing a tensor to a matrix, $\mathcal{M}(\cdot, \cdot, \cdot): \mathbb{R}_m^{n \times n \times\ldots n}\times [m] \times [m] \mapsto \mathbb{R}_2^{d \times d}$ as $$\mathcal{M}(\mathbf{A},1,2) = A_{j_1,j_2}= \sum_{j_3,j_4,\ldots, j_m=1}^d \mathbf{A}_{j_1,j_2,j_3,j_4,\ldots, j_m}.$$
\end{define}

\section{Hypergraph Block Models and Hypergraphons}
In this section, we introduce the hypergraphon model that we use for modeling higher order interactions. Before we do so, we introduce some basic definitions. 
\begin{define}[Hypergraph]
A hypergraph $G = (V,E)$ consists of a set of vertices denoted by a set and labeled as $V = [n]$ and a set of hyperedges $e \in E$ where each hyperedge $e$ consists of a subsets of vertices from $V$. A hypergraph is said to be $m$-uniform if every edge consists of exactly $m$ vertices. Note that when $m=2$ this corresponds to a graph.
\end{define}
In this paper, we restrict ourselves to hypergraphs for which hyperedges contain only unique vertices. Furthermore, we consider the case of $m$-uniform hypergraphs, motivated by the applications mentioned in Section~\ref{sec:intro}. As mentioned previously, a model for pairwise or higher-order interaction is represented by assuming particular structure on the adjacency matrix or tensor, denoted by $\mathbf{A} \in \{0,1\}^{n \times n \times \cdots \times n}$. Each entry in this tensor represents the presence or absence of an hyperedge. 

\subsection{Stochastic Block Models}\label{sec:sbm}
The stochastic block model in the graph setting was proposed by~\cite{holland1983stochastic} and has been analyzed extensively in the statistics~\citep{rohe2011spectral, lei2015consistency, choi2012stochastic, chatterjee2015matrix, abbe2017entrywise, Lesparse2017, gao2018community}, computer science~\citep{abbe2015community}, statistical physics~\citep{fortunato2010community, krzakala2013spectral} communities, thereby establishing a wide variety of results illustration several interesting phenomenon. Due to the large amount of literature developed on the topic of stochastic block models in the recent years, we refer the reader to~\cite{abbe2017community} for an in-depth survey. As mentioned before, all the above works consider the case of pairwise interaction models. \textcolor{black}{We now introduce the stochastic block model. Let $z: [n] \mapsto [k]$ denote a mapping of the set of $n$ vertices to one of the $k$ communities which is unknown. Let $\mathcal{Z}_{n,k}$ denote the set of all possible mappings $z:  [n] \mapsto [k]$. Note that the cardinality of the set $\mathcal{Z}$ is $k^n$. Let $\mathbf{Q} \in [0,1]^{k \times k}$ be a symmetric matrix (i.e., a $2^{nd}$ order tensor),  representing the probabilities of connections between and within each of the $k$ communities. The $\mathbf{A}_{j_1j_2}$-th entry of the adjacency matrix is modeled as a Bernoulli random variable with mean parameter $\mathbf{\Theta}_{j_1j_2}$. In the stochastic block model, the matrix $\mathbf{\Theta}$ has a block structure and takes values from the matrix $\mathbf{Q}$ in a particular fashion that we describe now. Specifically, the matrix $\mathbf{\Theta}$ is assumed to come from the following set $\mathcal{T}_z$ of matrices:
\begin{align}
\mathcal T_z = \{ \mathbf{\Theta} : \exists \mathbf{Q} \in [0,1]^{k \times k}~\&~z\in \mathcal{Z}_{n,k}~&\text{S.T.}~\mathbf{\Theta}_{j_1,j_2} = \mathbf{Q}_{a,b} = \mathbf{Q}_{b,a} \text{ for } (j_1,j_2) \in z^{-1}(a) \times z^{-1}(b), \nonumber\\&\text{ and } \mathbf{\Theta}_{j_1,j_1} = 0~\text{otherwise} \}. \nonumber
\end{align}}

Recently several authors have considered the straight-forward extension of the stochastic block model from the graph setting to the hypergraph setting. See for example~\cite{ghoshdastidar2015consistency, ke2019community, ahn2018hypergraph, pal2019community} for statistical and computational results in this setting. Furthermore, fundamental limits, in both estimation and hypothesis testing context was studied in~\cite{angelini2015spectral,chien2018community, kim2018stochastic, ahn2019community}.  Below, we describe the stochastic hypergraph block models, of which the standard stochastic block models are a special case. Let $\mathbf{Q} \in [0,1]^{k \times k \ldots \times k}$ be an $m^{th}$ order tensor that is assumed to by symmetric, i.e., $\mathbf{Q}_j = \mathbf{Q}_{\sigma(j)}$ for $\sigma \in \mathfrak{S}_m$. This tensor has the probabilities of connections between and within each of the $k$ communities. The $\mathbf{A}_j$-th entry of the tensor  is modeled as a Bernoulli random variable with mean parameter $\mathbf{\Theta}_{j}$. In the stochastic block model, the tensor $\mathbf{\Theta}$ has a block structure and takes values from the tensor $\mathbf{Q}$ in a particular fashion that we describe now. In this work we do not consider hyperedges which have at least one repeated vertex. Define $G \subset [n]^m$ to be the set of hyperedges with no repeated vertices. That is, $j \not\in G$ iff $\exists~ u,v$ s.t $j_u = j_v$. Then the tensor $\mathbf{\Theta}$ is assumed to come from the following set $\mathcal{T}_z$ of order $m$ tensors:
\begin{align}\label{eq:settz}
\mathcal T_z = \{ \mathbf{\Theta} : \exists \mathbf{Q} \in [0,1]_m^{k \times \cdots \times k}~\&~z\in \mathcal{Z}_{n,k}~\text{S.T.}~\mathbf{\Theta}_j = \mathbf{Q}_{z(j)} \text{ for } j \in G, \text{ and } \mathbf{\Theta}_j = 0~\text{otherwise} \}
\end{align}
Note that when $m=2$, the above definition corresponds to the case of standard stochastic block models defined for the case of graphs~\citep{holland1983stochastic} and described above.

\subsection{Hypergraphons}\label{sec:graphon}
The stochastic block model, as mentioned before serves as an intriguing model for understanding several fundamental principles behind statistical analysis of networks. Motivated by the theory of large limits~\citep{lovasz2012large}, recently nonparametric models and estimators for such models, was proposed and analyzed in~\cite{bickel2009nonparametric, airoldi2013stochastic, wolfe2013nonparametric, yang2014nonparametric, chan2014consistent, borgs2015consistent,  borgs2015private, gao2015rate, klopp2015oracle, zhang2017estimating, pensky2019dynamic}, for modeling pairwise interactions. The graphon models are in-homogenous random graph models that contain several specific models proposed in the literature, including the stochastic block model and latent space model~\citep{hoff2002latent}, as special cases. 
\begin{define}[Graphon and Sampling]\label{def:graphon}
Graphons are symmetric measurable functions $f: [0,1]^2 \mapsto [0,1]$. Given a graphon, the process of sampling a random graph proceeds as follows: 
\begin{enumerate}
\item A sequence of real numbers $X_1, \ldots, X_n$ is generated independently and uniformly on $[0,1]$. 
\item Let $\theta_{i,j} = f(X_i,X_j)$, where $f$ is a graphon.
\item Let $A_{i,j}$ be the entry of identity adjacency matrix. It is generated as $A_{i,j} =\textsc{ber}(\theta_{i,j})$.
\end{enumerate}
Note that topology of the network generated as above is invariant to the permuting the labels. Hence, we have a graph generated as above with a graphon $f(u,v)$ and another graphon $f(\vartheta(u), \vartheta(v))$ for a measure preserving bijection $\vartheta: [0,1] \mapsto [0,1]$ invoke the same random graph model.
\end{define}
We refer the reader to~\cite{lovasz2012large} for a complete and rigorous characterization of graphons. From the statistical literature, a model of particular relevance to us is that of sparse graphons as considered in~\cite{klopp2015oracle}. Under that model, step 2 in the above sampling procedure is modified as $\theta_{i,j} = \rho_n f(X_i,X_j)$, where $f$ is a graphon. Here $\rho_n > 0$ and it tends to zero as $\rho_n \to 0$ when $n \to \infty$. This provides a control on the number of edges generated in the random graph model. Specifically, if $\rho_n =1$ and fixed, then the expected number of edges is proportional to $n^2$ which corresponds to a dense graph. But for a value of $\rho_n \to 0$, as $n \to \infty$, we have the expected number of edge to be of the oder $O(\rho_n n^2)$. Hence by controlling the rate of decay of $\rho_n$, one can obtain sparse graphs that mimic real-world networks.  In particular, this modification is much needed when we later work with the hypergraphon setting as the number of higher order interactions are typically sparse as the order increases. \textcolor{black}{We would also like to mention that there are other approaches for sampling sparse exchangeable graph; we refer the intersted reader to~\cite{caron2017sparse, caron2017sparsity, veitch2019sampling, borgs2019sampling}.}

The study of the limit of hypergraphs and regularity and left-convergence results was initiated in~\cite{gowers2007hypergraph}. Several authors extended the results with most relevant ones being~\cite{elek2012measure} and~\cite{zhao2015hypergraph}. Below we first provide the definition of the hypergraphon. We introduce some new notations before we proceed. For a set $[m]$, denote by $\llbracket m \rrbracket$, the collection of all nonempty proper subsets of $[m]$ or equivalently the collection of nonempty subset of $[m]$ of size at most $m-1$. 

\begin{define}[$m$-uniform Hypergraphon and Sampling]\label{def:hypergraphon}
A $m$-uniform symmetric hypergraphon is a $2^m-2$ dimensional measurable function $f: [0,1]^{\llbracket m \rrbracket} \mapsto [0,1]$ whose coordinates are indexed by proper and non-empty proper subset of $[m]$. It is also assumed to be symmetric in the following sense: it remains invariant under any permutation of the coordinates induced by any permutation of $[m]$. Similar to the graphon model, we have the hypergraph generated as:
 \begin{enumerate}
\item A vector $X \in [0,1]^{\llbracket n \rrbracket}$  is generated uniformly randomly.
\item For a $j \in [n]^m$, let $\mathbf{\Theta}_j = f(X_j)$, where $f$ is graphon satisfying the conditions above and $X_j$ denotes co-ordinates of $X$ indexed by nonempty proper subsets of the set $j$. 
\item Let $\mathbf{A}_j$ be the entry of identity adjacency tensor, where $j$ contains only unique vertices. It is generated as $\mathbf{A}_j =\textsc{ber}(\mathbf{\Theta}_j)$. All other entries are set to zero.
\end{enumerate}
\end{define}
Note that the definition of symmetry in the above model is slightly different from the more standard notion of symmetry. It is best illustrated through an example. Consider $m=3$. In this case the hypergraphon is a $6$-dimensional function indexed by the sets $\big\{ \{1\}, \{2\}, \{3\}, \{1,2\}, \{2,3\}, \{1,3\} \big\}$. In this case, the function is invariant for any permutation $\sigma \in \mathfrak{S}_3$ operating on $[3]$. That is, it is invariant for any permutation of the first three coordinates and the last three co-ordinates.  Finally, similar to the case of sparse graphons we replace the step 2 in the above definition with $\mathbf{\Theta}_j = \rho_n f(X)$, where $f$ is a hypergraphon. Here $\rho_n > 0$ and it tends to zero as $\rho_n \to 0$ when $n \to \infty$ and hence provides a control on the number of hyperedges generated. Specifically the expected number of hyperedges generated under this model is of the order $O(\rho_n n^m)$. 

\subsection{The Class of  Smooth Lipschitz Hypergraphons}\label{sec:slh}
The difference between the graphons and hypergraphons is clear from the above Definition~\ref{def:graphon} and \ref{def:hypergraphon}. While graphons, which corresponds to the order-2 adjacency tensors  (or the standard adjacency matrices) are two dimensional functions, hypergraphons which corresponds to order-m adjacency tensors are $2^m-2$ dimensional functions and not $m$-dimensional functions as one might expect it to be. \textcolor{black}{The need for these additional dimensions could be understood by considering the two cases of generating $3$-uniform hypergraph: in the first case  each hyperedge (triangle) is sampled independently with same probability (1/2), and in the second case first a standard Erd\H{o}s-Renyi graphs on $n$ vertices is generated and then the 3-uniform hypergraph is formed based on the triangles in the graph. The limit of the former case is the constant (1/2) hypergraphon function. Where as, the limit of the later is different from the constant hypergraphon function and requires the additional coordinates to be represented. See~\cite{zhao2015hypergraph} or the example in Figure~\ref{fig:mishypergraphon} for the form of the limit in the later case. }



The class of general hypergraphons suffers from the problem of efficient estimation as the number of nodes requires must scale at a rate doubly exponentially in terms of the parameter $m$ for it to be consistent. Indeed, firstly the dimensionality of the graphon grows exponentially in terms of the cardinality of the hyperedges and secondly, the number of samples (in this case the nodes) required to achieve consistent estimation of an $d$ dimensional function already grows exponentially in $d$ (note here $d= 2^m$), as is well-known in nonparametric literature~\citep{klopp2015oracle, gine2015mathematical}. Furthermore, without smoothness assumptions, it might become more complicated to estimate such functions. In this section, we define the class of Simple Lipschitz Hypergraphon (SLH) which we will be main object we concentrate on in this paper.  

\begin{define}
A $m$-uniform Simple Lipschitz Hypergraphon, $\textsc{SLH}(m,L)$, is a measurable function $f: [0,1]^{m} \mapsto [0,1]$ whose coordinates are indexed by size-1 subsets of $[m]$. It is assumed to be symmetric in the following sense: it remains invariant under any permutation of the coordinates induced by any permutation of $[m]$. Furthermore, it is assumed to be $L$-Lipschitz. A consequence of the lipschitz assumption is that for any $(x_1, \ldots, x_m), (y_1,\ldots y_m) \in [0,1]^m $, we have 
\begin{align}\label{eq:smoothness}
| f(x_1, \ldots, x_m) - f (y_1,\ldots y_m) | \leq L \max_{i} |x_i - y_i|.
\end{align}
\end{define}

Note that the class of Lipschitz functions that we consider are a subset of $\alpha$-holder smooth functions ( with $\alpha =1$), which are standard function classes considered in the nonparametric regression literature (see for e.g.,~\citep{klopp2015oracle, gine2015mathematical}. Given the above class of functions, we follow the usual sampling procedure to generate a random hypergraph. The class of \emph{simple hypergraphons} was considered first in~\cite{kallenberg1999multivariate}, from the view point of modeling exchangeable arrays. The problem of estimation in graphons and hypergraphons was also considered~\cite{kallenberg1999multivariate} that provided asymptotic statements for the estimators. Furthermore, an interesting asymptotic result was proved in the same work on approximating general hypergraphons with simple hypergraphons, that provided precise statements on the quality of approximation. Motivated by this work, we restrict ourself to the case of simple hypergraphons as a tradeoff between modeling flexibility and efficient estimability. 

Once we restrict ourself to the class of SLH described above, one could use stochastic hypergraph models, with an appropriately growing number of blocks, to estimate hypergraphons efficiently. \textcolor{black}{Indeed, such an approach is motivated by the corresponding rate-optimal estimators proposed for the class of smooth graphons~\citep{gao2015rate, klopp2015oracle}.} While such estimators are suitable for estimation in simple hypergraphons, it is not suited for the case of general hypergraphons. This also highlights the limitations of block models in the modeling higher order interactions -- whereas general hypergraphons have much higher modeling capacity, stochastic block models with growing number of classes are suitable to capture only the case of SLH. 

 \textcolor{black}{We end this section by comparing the hypergraphon approach to several other approaches existing in the literature for modeling hypergraphs. Recall that~\cite{stasi2014beta} proposed a model or random hypergraphs with a given degree sequence, extending the work of~\cite{chatterjee2011random} who proposed a similar model for graphs. Such models are called as $\beta$-models in the literature. In~\cite{chatterjee2011random}, it was shown that in the case of graphs, the limit of such models could be characterized as a specific graphon function. Following a similar argument in~\cite{chatterjee2011random}, it is possible to show that the $\beta$-model for hypergraphs proposed in~\cite{stasi2014beta} also has a limiting hypergraphon construction, and characterize it precisely. It is interesting to examine and characterize the limiting hypergraphon corresponding to the approaches proposed by~\cite{ng2018model} and~\cite{turnbull2019latent}. This, however is beyond the scope of current work. A details study of this problem will be done in a future work.}









\section{Estimator and Main Results}\label{sec:mainresult}
We now introduce our estimator and establish rates of convergence for estimating the probability tensor, under the hypergraphon model. Our proof involves first establishing  the result when the probability tensor is generated by a block model and then showing that as we increase the number of blocks any function in the class SLH could be well approximated by the block model. Our proof technique extends the results of~\cite{klopp2015oracle} that provided a similar result for the case of graphons. In order do so, we extend their results with appropriate modifications to the case of simple hypergraphons. We also discuss the consequences of our result -- specifically the scaling with respect to the problem parameters like smoothness and order of the hypergraphon. 

As mentioned above, our estimator is based on a stochastic block model approximation to the class of SLH.  Let $\mathbf{\bar{\Theta}}$ denote the true probability tensor generated by a fixed hypergrahon $f_0$ from the class $SLH(m, L)$. Our estimator $\hat{\mathbf{\Theta}}$, is defined as the following  two-steps. In the first step, we solve a least-squares optimization problem. In the next step, we use the estimated assignment function and the probabilities to provide a final estimator of $\hat{\mathbf{\Theta}}$. They are summarized as follows:
\begin{alignat}{2}\label{ref:estimator}
\textbf{Step 1:}& \left(\hat z, \mathbf{\hat Q}\right) &&= \argmin_{z \in \mathcal{Z}_{n,k,n_0}; ~Q \in [0,1]_m^{ k \times \ldots \times k}} L(z,\mathbf{Q}),\\\nonumber
\textbf{Step 2:}&~~ ~~~~\hat{\mathbf{\Theta}}_j &&= ~~~\mathbf{\hat Q}_{\hat z(j)}.
\end{alignat}
Here, the loss function, $L(z,\mathbf{Q})$ is given by $$L(z,\mathbf{Q})= \sum_{a \in [k]^m} \sum_{ j \in z^{-1}(a) \cap H} \left( \mathbf{A}_j - \mathbf{Q}_{a} \right)^2$$ where $\mathcal{Z}_{n,k,n_0}$ is a slight modification of the set $\mathcal{Z}_{n,k}$ defined in Section~\ref{sec:sbm} -- it is defined as the set of all mappings from $[n]$ to $[k]$ such that $\min_{a \in [k]} |z^{-1}(a)| \geq n_0$. This modification is made to consider stochastic hypergraph models with \emph{balanced blocks}, while approximating the hypergraphon. For the above estimator, we have the following theorem that quantifies the rate of convergence of estimating the probability tensor $\mathbf{\bar{\Theta}}$.



\begin{thm}[Main Result] \label{thm:main}
Let $f_0 \in \textsc{SLH}(m, L)$, where $ L > 0$ and $0 < \rho_n \leq 1$. Let $k = \lceil (\rho_nn^m)^{\frac{1}{m+2}} \rceil$ and $n_0 \geq n/k$ and suppose $\rho_n \geq (\log n)^2 n^{-\frac{2}{m+1}}$. Then there is a constant $C$ depending only on $\alpha$ and $L$ such that the least squares estimate $\mathbf{\hat \Theta}$ constructed with this choice of $n_0$ satisfies

\[
	\E_{\mathbf{A}, X} \left[ \frac{1}{n^m} \| \hat{\mathbf{\Theta}} - \mathbf{\bar{\Theta}} \|_F^2 \right] \leq C \left\{ \rho_n^{\frac{2m + 2}{m + 2}} n^{-\frac{2m}{m+2}} +  \rho_n \left( \frac{\log n}{n^{m-1}} \right) \right\}.
\]

\end{thm}

\begin{proof}[Proof of Theorem~\ref{thm:main}]
As mentioned before, our strategy is to approximate the hypergraphon from the class $SLH(m, L)$ by a hypergraph stochastic block model with appropriately defined number of communities $k$. Given such a construction, we then capture the estimation error of parameter estimation in the hypergraph stochastic block model. Let $\mathbf{\Theta}_*$ be the best $k$-class hypergraph stochastic block model approximation of $\mathbf{\bar{\Theta}}$ in Frobenius norm from the set,
\begin{align*}
\mathcal T_{z,n_0} = \{ \mathbf{\Theta} : \exists \mathbf{Q} \in [0,1]_m^{k \times \cdots \times k}~\&~z\in \mathcal{Z}_{n,k,n_0}~\text{S.T.}~\mathbf{\Theta}_j = Q_{z(j)} \text{ for } j \in G, \text{ and } \mathbf{\Theta}_j = 0~\text{otherwise} \}.
\end{align*}

Note that the set $\mathcal T_{z,n_0}$ is a slight modification of the set $\mathcal T_{z}$ defined in Equation~\ref{eq:settz} to enforce the balanced partition constraint via choosing $n_0$. Based on this, we first have the following lemma, that decomposes the overall error into an estimation error term and an approximation error term.
\begin{lemma}[Estimation Error]\label{lem:est} 
In the hypergraph stochastic block model setting, we can find an estimator $\hat{\mathbf{\Theta}}$ such that there exists an absolute positive constant $C_1 > 0$ and positive constants $C_2, C_3 > 0$ depending on $m$ such that for $n_0 \geq 2$,
\[
	\E_{\mathbf{A},X}\left[ \frac1{n^m} \| \hat{\mathbf{\mathbf{\Theta}}} - \mathbf{\bar{\Theta}} \|_F^2 \right] \leq \frac{C_1}{n^m}\E_{X}\left[  \| \mathbf{\bar{\Theta}} - \mathbf{\Theta}_* \|_F^2\right] + C_2 \| \mathbf{\bar{\Theta}} \|_\infty \left( \frac{\log k}{n^{m-1}} + \frac{k^m}{n^m} \right) + \frac{C_3 \log n/n_0}{n_0} \left( \frac{\log k}{n^{m-1}} + \frac{k^m}{n^m} \right)
\]
\end{lemma}
Note that our assumption $\rho_n \geq (\log n)^2 n^{-\frac{2}{m+1}}$, after some algebra gives us $\rho_n \geq \frac{\log n/n_0}{n_0}$. 
Under this scaling, note that third term in the expectation bound of Lemma~\ref{lem:est} could be absorbed in the second term and hence we get
\[
	\E_{\mathbf{A},X} \left[ \frac{1}{n^m} \| \hat{\mathbf{\Theta}} - \mathbf{\bar{\Theta}} \|_F^2 \right] \leq \frac{C}{n^m}\E_{X}\left[  \| \mathbf{\bar{\Theta}} - \mathbf{\Theta}_* \|_F^2 \right]+ C\rho_n \left( \frac{\log k}{n^{m-1}} + \frac{k^m}{n^m} \right)
\]
Here the term $\frac{1}{n^m}\E_{X}\left[  \| \mathbf{\bar{\Theta}} - \mathbf{\Theta}_* \|_F^2\right]$, corresponds to the approximation error part. We now have the following lemma that bounds the approximation error.
\begin{lemma}[Approximation Error]\label{lem:approx}
Consider the hypergraphon model with $f_0 \in \textsc{SLH}(m, L)$ where $L > 0$. Let $n_0 \geq 2$ and $k = \lfloor n/n_0 \rfloor$. Then,
\[
	\E_X\left[ \frac{1}{n^m} \| \mathbf{\bar{\Theta}} - \mathbf{\Theta}_{*} \|_F^2 \right] \leq C M^2 \rho_n^2 \left( \frac{1}{k^2} \right)
\]
\end{lemma}

By Lemma \ref{lem:approx}, our bound on approximation error, we get the following bound for the overall error:

\[
	\E_{\mathbf{A},X} \left[ \frac{1}{n^m} \| \hat{\mathbf{\Theta}} - \mathbf{\bar{\Theta}} \|_F^2 \right] \leq C \left\{ \frac{\rho_n^2}{k^{2}} + \rho_n \left( \frac{\log k}{n^{m-1}} + \frac{k^m}{n^m} \right) \right\}.
\]

We now choose $k = \lceil (\rho_nn^m)^{\frac{1}{m+2}} \rceil$ with the goal of balancing the first and third terms in the above expression. This choice of $k$ gives us
\[
	\E_{\mathbf{A},X}\left[  \frac{1}{n^m} \| \hat{\mathbf{\Theta}} - \mathbf{\bar{\Theta}} \|_F^2 \right] \leq C \left\{ \rho_n^{\frac{2m + 2}{m + 2}} n^{-\frac{2m}{m+2}} +  \rho_n \left( \frac{\log n}{n^{m-1}} \right) \right\},
\]
which completes the proof of Theorem~\ref{thm:main}. The proofs of Lemma~\ref{lem:est} and \ref{lem:approx} are more involved and are provided in the Appendix~\ref{sec:supp}.
\end{proof}




\begin{remark}
Note that for our result, we require the sparsity parameter $\rho_n > n^{-2/(m+1)}$. This corresponds to moderately sparse regime. Extending the result to the case of highly sparse regime (i.e., $\rho_n > n^{-(m-1)}$) is an interesting problem. It appears that the current proof technique is incapable of handling the highly sparse regime and a completely different approximation of the hypergraphon (for example, a multi-level approximation) with a parametric model might be required. This highlights another important distinction between the graphons and hypergraphons. We leave this interesting problem as future work.
\end{remark}
\begin{remark}
Our result above could be extended to the case of $\alpha$-Holder simple hypergraphons in a straight forward manner. In that case, there are two regimes in which our estimator behaves differently. For simplicity, consider the case when $\rho_n =1$. In this case, we have
\[ \E_{\mathbf{A},X}\left[ \frac{1}{n^m} \| \hat{\mathbf{\Theta}} - \mathbf{\bar{\Theta}} \|_F^2 \right] =
  \begin{cases}
    n^{-\frac{2\alpha'm}{m+2\alpha}}    & \quad \text{when } 0\leq \alpha < 1\\
    n^{-\frac{2m}{m+2}} & \quad \text{when} ~\alpha \geq  1 .
  \end{cases}
\]
\end{remark}
\noindent In particular, for no value of $\alpha$, the second term $(\frac{\log n}{n^{m-1}})$ becomes the dominant term. Furthermore, when $m=2$, which corresponds to the case of graphons, for $\alpha\geq 1$, the first term and the second term are equivalent up to $\log$ factors and we recover the results of~\cite{klopp2015oracle}.  We conjecture that our rates are essentially minimax optimal for the SLH when $\rho=1$. In order to see that note that the problem of estimating a hypergraphon from the class SLH corresponds to the case of estimating a $m$-dimensional Lipschitz function. The minimax rate of non-parametric regression in this case essentially coincides with the rates we obtained (see for e.g.,~\cite{gine2015mathematical} for a comprehensive overview of nonparametric models in the context of regression and density estimation). It is interesting to examine the minimax optimality of our result for the general case of $\rho_n$. A proof of minimax rates for general $\rho$ has eluded us thus far.


\section{Algorithm}\label{sec:algorithm}
Recall that our estimator $\hat{\mathbf{\Theta}}$, involves solving the following least-squares optimization problem
\[
	\hat z, \mathbf{\hat Q} = \argmin_{z, Q} L(z,\mathbf{Q}),\quad
\text{where}\quad
	L(z,\mathbf{Q}) = \sum_{j \in [n]^m} \left( \mathbf{A}_j - \mathbf{Q}_{z(j)} \right)^2.
\]
An immediate algorithm to optimize the above objective function is to perform coordinate descent on the above objective function. This approach has the following drawback. While for a fixed $z$ optimizing for $\mathbf{Q}$ is a standard least-squared problem, for a fixed $\mathbf{Q}$ optimizing for $z$ evaluating all possible assignment functions $z \in \mathcal Z_{n,k}$, which would take exponential time and would thus be computationally infeasible even for moderate values of $n, k$. One could further break optimizing for $z$ into optimizing for the individual coordinates $z(j)$. But we found through preliminary experiments such an approach runs into several issues like non-convergence of the iterates and worse performance (when it converges). In order to obtain a computable estimator and demonstrate our results empirically, we propose to use the procedure described in Algorithm~\ref{alg:alternating}. 
\begin{algorithm}[!htbp]
\caption{Alternating Minimization for hypergraphon}
\begin{algorithmic}
 \STATE {\bfseries Input:} $\mathbf{A}$, $k$. 
 \vspace{0.05in}
 \REPEAT
\STATE Let $E_{ia}$ be the number of hyperedges containing both $i$ and a member of $a$ according to the current $\hat z$.
 \begin{align}\label{eq:eiaupdate}
 	E_{ia} =  \sum_{j_2 \in \hat{z}^{(-1)}(a)} \sum_{j_3,\cdots,j_m \in [n]} \mathbf{A}_{ij_2\cdots j_m}
 \end{align}
 \STATE Update $\hat z$ as
 \[
 	\hat z(i) =  \argmax_a \frac{1}{\varkappa_a} E_{ia}
\]
where $\varkappa_a$ is defined in Equation~\ref{eq:temp}.
\UNTIL{Convergence} 
\vspace{0.05in}
\STATE \textbf{Compute} $	 \mathbf{\hat Q}_a = \frac{1}{\eta_a} \sum_{j \in \hat z^{(-1)}(a)} A_j
$
\vspace{0.05in}
\STATE {\bfseries Output:} $\mathbf{\hat Q}_a$ and $\hat z$.
\end{algorithmic} \label{alg:alternating} 
\end{algorithm}

\noindent \textbf{Intuition}: We now describe the main intuition behind the algorithm. Firstly note that when $a \in [k]$, $\eta_a$ is the number of vertices which are assigned community $a$, and when $a \in [k]^m$, $\eta_a$ is the number of hyperedges whose community assignments match $a$ node-wise. In other words, $j \in [n]^m$ matches $a \in [k]^m$ if there exists a permutation $\sigma \in \mathfrak{S}_k$, such that for all $i \in [n]$, $a_i = \sigma(z(j_i))$. With this observation, for a fixed $z$, we have $\mathbf{Q}_{a} = \frac{1}{\eta_\alpha} \sum_{j \in z^{-1}(a) }\mathbf{A}_{j}$. Furthermore for any minimizer $\hat{\mathbf{Q}}$ and $\hat z$ of $L(\mathbf{Q},z)$ we have $$\hat{\mathbf{Q}}_a = \frac{1}{\eta_\alpha} \sum_{j \in \hat z^{-1}(a) }\mathbf{A}_{j}.$$ Hence we concentrate first on estimate $\hat z$ and then use the above relation to obtain an estimate for $\hat{\mathbf{Q}}$ and subsequently $\hat{ \mathbf{\Theta}}$. 

The procedure in Algorithm~\ref{alg:alternating} is motivated by the modified version of k-means type algorithm analyzed in~\cite{lu2016statistical} for community detection in standard stochastic block model. Their algorithm repeatedly updates $z(i)$ according to the \emph{fraction} of nodes in each community $a$ that node $i$ connects to. This can be viewed as the fraction of possible edges into $a$ which $i$ is a part of. But this intuition does not extend straight-forwardly to the case of hypergraphs. In order to extend their approach to the case of hypergraphs, we first propose to collapse the $m$-th order adjacency tensor $\mathbf{A}$ to the matrix $\mathcal{M}(\mathbf{A})$ (recall the definition from section~\ref{sec:notation}). Now we define $E_{ia}$, the number of hyperedges containing for $i$ and a member of $a$, according to the Equation~\ref{eq:eiaupdate}. Note that is not exactly equal to the number of hyperedges containing for $i$ and a member of $a$, as we are overcounting hyperedges which contain more than one element assigned community $a$. Even with this discrepancy, our algorithm performs well empirically as we demonstrate in section~\ref{sec:simexp}. This matrix is then used to estimate $z(i)$. Our update for $z(i)$ is based on the fraction of possible hyperedges that node $i$ is a part of. The total number of possible hyperedges containing both $i$ and a member of community $a$ is given by 
\begin{align}\label{eq:temp}
\varkappa_a=	\binom{\eta_a}{1}\binom{n-\eta_a}{m-2} + 2!\binom{\eta_a}{2}\binom{n-\eta_a}{m-3} + \ldots + (m-1)!\binom{\eta_a}{m-1}\binom{n-\eta_a}{0},
\end{align}
where we count a hyperedge $j$ times if it contains $j$ elements from $a$, to match the overcounting of $E_{ia}$.\\

\noindent \textbf{Initialization}: Like most alternating methods, our algorithm is susceptible to local minima. To overcome this issue, we use two methods. First, we try many random initializations, running the algorithm to convergence, and then selecting the best result according to our empirical loss function $L(z,Q)$. Our second method is to initialize $z$ using a spectral classifier, as given by~\cite{ghoshdastidar2016uniform}. Their method is again based on the performing spectral clustering on the $\mathcal{M}(\mathbf{A})$ matrix, as this operation preserves the factors approximately. We report the results using both initialization methods in section~\ref{sec:simexp} and discuss the similarities and differences between them.  

We also note that the tensor collapse operation also bears similarities with the Leurgan's algorithm for decomposing tensors~\citep{leurgans1993decomposition}. The algorithm, proposed specifically for the case of three-way tensors, involves taking weighted sums of the slices. Note that in our case, there is no need for taking a weighted sum. Finally, the theoretical analysis of this algorithm is more involved than the standard graph based stochastic block model case. We plan to report the theoretical results in the context of community detection in hypergraphs in the near future.



\section{Simulation Results: Well-specified Case}\label{sec:simexp}
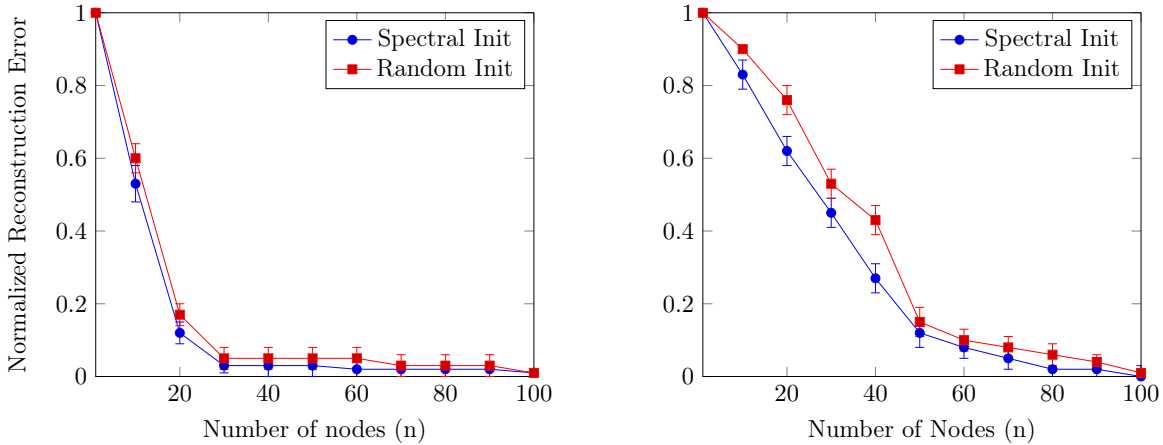
\begin{figure*}[t]
\centering
\begin{minipage}{0.5\textwidth}
\centering
\begin{tikzpicture}[scale=0.85]
  \begin{axis}[
    xlabel = $\text{Number of nodes (n)}$,
ylabel=$\text{Normalized Reconstruction Error}$,
xmax = 100,
xmin = 1,
ymax = 1,
ymin = 0
]
\addplot+[error bars/.cd,
y dir=both,y explicit]
 coordinates {
( 1, 1 )+- (0.0, 0.007)
( 10, 0.53 )+- (0.0, 0.05)
( 20, 0.12 )+- (0.0, 0.03)
( 30, 0.03 )+- (0.0, 0.02)
( 40, 0.03 )+- (0.0, 0.03)
( 50, 0.03 )+- (0.0, 0.03)
( 60, 0.02 )+- (0.0, 0.03)
( 70, 0.02 )+- (0.0, 0.02)
( 80, 0.02 )+- (0.0, 0.02)
( 90, 0.02 )+- (0.0, 0.02)
( 100, 0.01 )+- (0.0, 0.00)
};  \addlegendentry{Spectral Init} ;
\addplot+[error bars/.cd,
y dir=both,y explicit]
 coordinates {
( 1, 1 )+- (0.0, 0.007)
( 10, 0.60 )+- (0.0, 0.04)
( 20, 0.17 )+- (0.0, 0.03)
( 30, 0.05 )+- (0.0, 0.03)
( 40, 0.05 )+- (0.0, 0.03)
( 50, 0.05 )+- (0.0, 0.03)
( 60, 0.05 )+- (0.0, 0.03)
( 70, 0.03 )+- (0.0, 0.03)
( 80, 0.03 )+- (0.0, 0.03)
( 90, 0.03 )+- (0.0, 0.03)
( 100, 0.01 )+- (0.0, 0.00)
};  \addlegendentry{Random Init} ;
\end{axis}
\end{tikzpicture}
\end{minipage}
\begin{minipage}{0.5\textwidth}
\centering
\begin{tikzpicture}[scale=0.85]
  \begin{axis}[
  xlabel = $\text{Number of Nodes (n)}$,
ylabel=$$,
xmax = 100,
xmin = 1,
ymax = 1,
ymin = 0
]
\addplot+[error bars/.cd,
y dir=both,y explicit]
 coordinates {
( 1, 1 )+- (0.0, 0.007)
( 10, 0.83 )+- (0.0, 0.04)
( 20, 0.62 )+- (0.0, 0.04)
( 30, 0.450 )+- (0.0, 0.04)
( 40, 0.270 )+- (0.0, 0.04)
( 50, 0.12 )+- (0.0, 0.04)
( 60, 0.08 )+- (0.0, 0.03)
( 70, 0.05 )+- (0.0, 0.03)
( 80, 0.02 )+- (0.0, 0.03)
( 90, 0.02 )+- (0.0, 0.03)
( 100, 0.00 )+- (0.0, 0.03)
};  \addlegendentry{Spectral Init} ;

\addplot+[error bars/.cd,
y dir=both,y explicit]
 coordinates {
( 1, 1 )+- (0.0, 0.007)
( 10, 0.9 )+- (0.0, 0.005)
( 20, 0.76 )+- (0.0, 0.04)
( 30, 0.53 )+- (0.0, 0.04)
( 40, 0.43 )+- (0.0, 0.04)
( 50, 0.15 )+- (0.0, 0.04)
( 60, 0.10 )+- (0.0, 0.03)
( 70, 0.08 )+- (0.0, 0.03)
( 80, 0.06 )+- (0.0, 0.03)
( 90, 0.04 )+- (0.0, 0.02)
( 100, 0.01 )+- (0.0, 0.00)
};  \addlegendentry{Random Init} ;
\end{axis}
\end{tikzpicture}
\end{minipage}\hfill
\caption{The graphs correspond to normalized average estimation error, for estimating the probability matrix, as a function of the number of nodes. The left hand side corresponds to when the hypergraph was sampled from hypergraphon in case 1 and the right hand side corresponds to the case 2. The value of $\rho_n$ was set to $1$ and hence this corresponds to the \textbf{dense} edges situation. Each point in the graphs above corresponds to a average over 50 independent trials and the bars represent the corresponding standard error. The value of $k$ was fixed at  $0.6\cdot n^3$ (based on the insight provided by the theorem). }
\label{fig:wellspecifieddense}
\end{figure*}

We now provide simulation results depicting the performance of our algorithm for the case when the random graphs are generated from the following two simple hypergraphons:
\begin{align*}
\text{Case 1:}~f(u,v,w) &= uvw \\
\text{Case 2:}~f(u,v,w) &= \frac{1}{1+e^{\{-(c_1 u^2+c_2v^2+c_3w^2)\}}}
\end{align*}
These experiments corresponds to well-specified modeling situation -- we assumed the true graphon is a simple hypergraphon and we use the proposed algorithm to estimate the probability matrix. In Section~\ref{sec:misspec}, we also consider the misspecified case -- here we test how well the algorithm performs for the case when the true hypergraphon is not a simple hypergraphon. \textcolor{black}{Our error metric is the normalized $L_2$ reconstruction error (i.e., $ \| \hat{\mathbf{\Theta}} - \mathbf{\bar{\Theta}} \|_F^2/ \|  \mathbf{\bar{\Theta}} \|_F^2$)}. We use both the random and spectral initialization described in previous section. \textcolor{black}{Our results for the well-specified case are reported in Figures~\ref{fig:wellspecifieddense} and~\ref{fig:wellspecifiedsparse}. For each case, we performed the simulation for 50 independent trials and the average values are reported. The bars in the figures represent the corresponding standard error. The value of $\rho_n$ was set to $1$ and $0.7$ in Figure~\ref{fig:wellspecifieddense} and~\ref{fig:wellspecifiedsparse} respectively, to correspond to the \textbf{dense} edges setting and \textbf{moderately sparse} edges situation. The value of $k$ was fixed at  $0.6\cdot n^3$ and $0.5\cdot n^3$ in Figure~\ref{fig:wellspecifieddense} and~\ref{fig:wellspecifiedsparse} respectively, based on the insight provided by the theorem. We experimented with both random initialization and spectral initialization as described in the previous section.} Below are our observations from the experiments.
\begin{itemize}
\item The main difference between the two hypergraphons is that the one in case 1 could be thought of as a rank-1 function. Since our algorithm is based on the \textsc{tensorcollapse} operation, the performance depends on how well such an operation captures the spectrum of the true function. It is easy to see that in the case of rank-1 function, the operation completely preserves the spectrum. Hence we expect our algorithm to perform well in case 1 than in case 2. Indeed this is reflect in Figures~\ref{fig:wellspecifieddense} and~\ref{fig:wellspecifiedsparse} for both the dense and sparse cases.
\item The difference between the two initialization schemes is nearly negligible in the dense setting but the spectral initialization performs comparatively better in the sparse setting. We believe with a more sophisticated regularized spectral initialization one could achieve better performance. Precisely characterizing this in a theoretical framework is left a future work. 
\item As expected, as the number of nodes increases the expected reconstruction error decreases confirming the theoretical result presented in Theorem~\ref{thm:main}. Note that the value of $k$ is set at a fixed value in the experiment. 
\end{itemize}

\subsection{Effect of number of blocks}
Recall that our algorithm takes in as input $k$, the number of communities in the hypergraph stochastic block model that is used to approximate the hypergraphon. In Figure~\ref{fig:effectofk} the results of increasing the number of communities is depicted. \textcolor{black}{The simulation setup is same as in Section~\ref{sec:simexp} for the case of dense graphs; that is, $\rho_n=1$. We considered $n=20$ so that we could clearly observe the bias-variance tradeoff.  For each case, we performed the simulation for 50 independent trials and the average values are reported. The bars in the figures represent the corresponding standard error.} We use hypergraphs samples from both case 1 and 2 hypergraphons and increase $k$ from $0.1\cdot n^3$ to $0.9\cdot n^3$ in steps of  $0.1$. The graphs in Figure~\ref{fig:effectofk} demonstrate a bias-variance tradeoff in terms of the effect of $k$ on estimating the probability matrix. For lower values of $k$, the corresponding block model does not provide a good approximation for the underlying simple hypergraphons while for higher values of $k$, the number of nodes in each community is too less to provide any information for estimating the probabilities of edge formulation within that community. There is a sweet-spot in the middle which achieves the optimal value of $k$ that balance the above bias and variance. This insight was used in the experiments presented in the previous section to fix the value of $k$.
  
\begin{figure*}[t]
\centering
\begin{minipage}{0.5\textwidth}
\centering
\begin{tikzpicture}[scale=0.85]
  \begin{axis}[
    xlabel = $\text{Number of nodes (n)}$,
ylabel=$\text{Normalized Reconstruction Error}$,
xmax = 100,
xmin = 1,
ymax = 1,
ymin = 0
]
\addplot+[error bars/.cd,
y dir=both,y explicit]
 coordinates {
( 1, 1 )+- (0.0, 0.007)
( 10, 0.57 )+- (0.0, 0.05)
( 20, 0.14 )+- (0.0, 0.03)
( 30, 0.04 )+- (0.0, 0.02)
( 40, 0.04 )+- (0.0, 0.03)
( 50, 0.03 )+- (0.0, 0.03)
( 60, 0.02 )+- (0.0, 0.03)
( 70, 0.02 )+- (0.0, 0.02)
( 80, 0.02 )+- (0.0, 0.02)
( 90, 0.02 )+- (0.0, 0.02)
( 100, 0.01 )+- (0.0, 0.00)
};  \addlegendentry{Spectral Init} ;
\addplot+[error bars/.cd,
y dir=both,y explicit]
 coordinates {
( 1, 1 )+- (0.0, 0.007)
( 10, 0.64 )+- (0.0, 0.04)
( 20, 0.18 )+- (0.0, 0.03)
( 30, 0.07 )+- (0.0, 0.03)
( 40, 0.06 )+- (0.0, 0.03)
( 50, 0.05 )+- (0.0, 0.03)
( 60, 0.05 )+- (0.0, 0.03)
( 70, 0.02 )+- (0.0, 0.03)
( 80, 0.03 )+- (0.0, 0.03)
( 90, 0.02 )+- (0.0, 0.03)
( 100, 0.01 )+- (0.0, 0.00)
};  \addlegendentry{Random Init} ;
\end{axis}
\end{tikzpicture}
\end{minipage}
\begin{minipage}{0.5\textwidth}
\centering
\begin{tikzpicture}[scale=0.85]
  \begin{axis}[
  xlabel = $\text{Number of Nodes (n)}$,
ylabel=$$,
xmax = 100,
xmin = 1,
ymax = 1,
ymin = 0
]
\addplot+[error bars/.cd,
y dir=both,y explicit]
 coordinates {
( 1, 1 )+- (0.0, 0.007)
( 10, 0.86 )+- (0.0, 0.04)
( 20, 0.60 )+- (0.0, 0.04)
( 30, 0.450 )+- (0.0, 0.04)
( 40, 0.280 )+- (0.0, 0.04)
( 50, 0.15 )+- (0.0, 0.04)
( 60, 0.06 )+- (0.0, 0.03)
( 70, 0.05 )+- (0.0, 0.03)
( 80, 0.02 )+- (0.0, 0.03)
( 90, 0.02 )+- (0.0, 0.03)
( 100, 0.00 )+- (0.0, 0.03)
};  \addlegendentry{Spectral Init} ;

\addplot+[error bars/.cd,
y dir=both,y explicit]
 coordinates {
( 1, 1 )+- (0.0, 0.007)
( 10, 0.93 )+- (0.0, 0.005)
( 20, 0.74 )+- (0.0, 0.04)
( 30, 0.51 )+- (0.0, 0.04)
( 40, 0.38 )+- (0.0, 0.04)
( 50, 0.12 )+- (0.0, 0.04)
( 60, 0.10 )+- (0.0, 0.03)
( 70, 0.06 )+- (0.0, 0.03)
( 80, 0.06 )+- (0.0, 0.03)
( 90, 0.04 )+- (0.0, 0.02)
( 100, 0.01 )+- (0.0, 0.00)
};  \addlegendentry{Random Init} ;
\end{axis}
\end{tikzpicture}
\end{minipage}
\caption{The graphs correspond to normalized average estimation error, for estimating the probability matrix, as a function of the number of nodes. The left hand side corresponds to when the hypergraph was sampled from hypergraphon in case 1 and the right hand side corresponds to the case 2. The value of $\rho_n$ was set to $0.7$ and hence this corresponds to the \textbf{sparse} edges situation. Each point in the graphs above corresponds to a average over 50 independent trials and the bars represent the corresponding standard error. The value of $k$ was fixed at  $0.5\cdot n^3$ (based on the insight provided by the theorem). }
\label{fig:wellspecifiedsparse}
\end{figure*}

\begin{figure*}[t]
\centering
\begin{minipage}{0.5\textwidth}
\centering
\begin{tikzpicture}[scale=0.85]
  \begin{axis}[
    xlabel = $\text{Value of $k$}$,
ylabel=$\text{Normalized Reconstruction Error}$,
xmax = 11,
xmin = 4,
ymax = 1,
ymin = 0
]
\addplot+[error bars/.cd,
y dir=both,y explicit]
 coordinates {
( 4, 0.76 )+- (0.0, 0.06)
( 5, 0.60 )+- (0.0, 0.05)
( 6, 0.48 )+- (0.0, 0.03)
( 7, 0.32 )+- (0.0, 0.04)
( 8, 0.19 )+- (0.0, 0.03)
( 9, 0.15 )+- (0.0, 0.03)
( 10, 0.20 )+- (0.0, 0.03)
( 11, 0.36 )+- (0.0, 0.03)
};  \addlegendentry{Spectral Init} ;
\addplot+[error bars/.cd,
y dir=both,y explicit]
 coordinates {
( 4 ,0.79 )+- (0.0, 0.06)
( 5, 0.63 )+- (0.0, 0.05)
( 6, 0.52 )+- (0.0, 0.03)
( 7, 0.35 )+- (0.0, 0.04)
( 8, 0.20 )+- (0.0, 0.03)
( 9, 0.17 )+- (0.0, 0.03)
( 10, 0.250 )+- (0.0, 0.03)
( 11, 0.370 )+- (0.0, 0.03)
};  \addlegendentry{Random Init} ;
\end{axis}
\end{tikzpicture}
\end{minipage}\hfill
\begin{minipage}{0.5\textwidth}
\centering
\begin{tikzpicture}[scale=0.85]
  \begin{axis}[
  xlabel = $\text{Value of $k$}$,
ylabel=$$,
xmax = 11,
xmin = 4,
ymax = 1,
ymin = 0
]
\addplot+[error bars/.cd,
y dir=both,y explicit]
 coordinates {
( 4, 0.75 )+- (0.0, 0.06)
( 5, 0.56 )+- (0.0, 0.05)
( 6, 0.47 )+- (0.0, 0.03)
( 7, 0.30 )+- (0.0, 0.04)
( 8, 0.19 )+- (0.0, 0.03)
( 9, 0.20 )+- (0.0, 0.03)
( 10, 0.22 )+- (0.0, 0.03)
( 11, 0.38 )+- (0.0, 0.03)
};  \addlegendentry{Spectral Init} ;
\addplot+[error bars/.cd,
y dir=both,y explicit]
 coordinates {
( 4 ,0.79 )+- (0.0, 0.06)
( 5, 0.63 )+- (0.0, 0.05)
( 6, 0.52 )+- (0.0, 0.03)
( 7, 0.35 )+- (0.0, 0.04)
( 8, 0.20 )+- (0.0, 0.03)
( 9, 0.21 )+- (0.0, 0.03)
( 10, 0.250 )+- (0.0, 0.03)
( 11, 0.370 )+- (0.0, 0.03)
};  \addlegendentry{Random Init} ;
\end{axis}
\end{tikzpicture}
\end{minipage}\hfill
\caption{The graphs correspond to normalized average estimation error, for estimating the probability matrix, as a function of increasing the number of blocks. The left hand side corresponds to when the hypergraph was sampled from hypergraphon in case 1 and the right hand side corresponds to the case 2. The value of $\rho_n$ was set to $1$ and hence this corresponds to the dense edges situation. Each point in the graphs above corresponds to a average over 50 independent trials and the bars represent the corresponding standard error. The $x$-axis is to be interpreted after scaling it with $ n^3$.}
\label{fig:effectofk}
\end{figure*}

\subsection{\textcolor{black}{Runtime comparison}}\label{sec:wallclock}
\textcolor{black}{Modeling higher-order interactions certainly comes at a computational cost compared to that of pairwise interactions. In this section, we compare the wall-clock times of random-initialization method for the above experiments. We consider random-initialization method as we observe that it performs as good as the one with spectral initialization. The run-times of the spectral-initialization method would take into account the time for obtaining the initialization, in addition.} 

\begin{figure*}[t]
\centering
\begin{minipage}{0.5\textwidth}
\centering
\begin{tikzpicture}[scale=0.85]
  \begin{axis}[
    xlabel = $\text{Number of nodes (n))}$,
legend pos=north west,
ylabel=$\text{Wall-clock time (in minutes)}$,
xmax = 100,
xmin = 10,
ymax = 7,
ymin = 0
]
\addplot+[error bars/.cd,
y dir=both,y explicit]
 coordinates {
( 10, 0.56 )+- (0.0, 0.4)
( 20, 1.0 )+- (0.0, 0.4)
( 30, 1.450 )+- (0.0, 0.4)
( 40, 2.0 )+- (0.0, 0.4)
( 50, 2.55 )+- (0.0, 0.4)
( 60, 3.06 )+- (0.0, 0.3)
( 70, 3.6 )+- (0.0, 0.3)
( 80, 4.02 )+- (0.0, 0.3)
( 90, 5.02 )+- (0.0, 0.3)
( 100, 6.80 )+- (0.0, 0.3)
};  \addlegendentry{Dense case 1} ;
\addplot+[error bars/.cd,
y dir=both,y explicit]
 coordinates {
( 10, 0.49 )+- (0.0, 0.4)
( 20, 1.1 )+- (0.0, 0.4)
( 30, 1.50 )+- (0.0, 0.4)
( 40, 2.10 )+- (0.0, 0.4)
( 50, 2.45 )+- (0.0, 0.4)
( 60, 3.16 )+- (0.0, 0.3)
( 70, 3.56 )+- (0.0, 0.3)
( 80, 4.12 )+- (0.0, 0.3)
( 90, 5.42 )+- (0.0, 0.3)
( 100, 6.90 )+- (0.0, 0.3)
};  \addlegendentry{Dense case 2} ;
\addplot+[error bars/.cd,
y dir=both,y explicit]
 coordinates {
( 10, 0.39 )+- (0.0, 0.4)
( 20, 0.95 )+- (0.0, 0.4)
( 30, 1.280 )+- (0.0, 0.4)
( 40, 1.70 )+- (0.0, 0.4)
( 50, 2.25 )+- (0.0, 0.4)
( 60, 3.16 )+- (0.0, 0.3)
( 70, 3.56 )+- (0.0, 0.3)
( 80, 4.32 )+- (0.0, 0.3)
( 90, 5.42 )+- (0.0, 0.3)
( 100, 6.80 )+- (0.0, 0.3)
};  \addlegendentry{Sparse case 1} ;
\addplot+[error bars/.cd,
y dir=both,y explicit]
 coordinates {
( 10, 0.29 )+- (0.0, 0.4)
( 20, 0.9 )+- (0.0, 0.4)
( 30, 1.30 )+- (0.0, 0.4)
( 40, 1.80 )+- (0.0, 0.4)
( 50, 2.25 )+- (0.0, 0.4)
( 60, 3.06 )+- (0.0, 0.3)
( 70, 3.36 )+- (0.0, 0.3)
( 80, 4.02 )+- (0.0, 0.3)
( 90, 5.22 )+- (0.0, 0.3)
( 100, 6.80 )+- (0.0, 0.3)
}; \addlegendentry{Sparse case 2} ;
\end{axis}
\end{tikzpicture}
\end{minipage}\hfill
\begin{minipage}{0.5\textwidth}
\centering
\begin{tikzpicture}[scale=0.85]
  \begin{axis}[
legend pos=north west,
  xlabel = $\text{Value of $k$}$,
ylabel=$$,
xmax = 11,
xmin = 4,
ymax = 5,
ymin = 0
]
\addplot+[error bars/.cd,
y dir=both,y explicit]
 coordinates {
( 4, 0.6 )+- (0.0, 0.6)
( 5, 1.1 )+- (0.0, 0.5)
( 6, 1.7 )+- (0.0, 0.3)
( 7, 2.10 )+- (0.0, 0.4)
( 8, 2.7 )+- (0.0, 0.3)
( 9, 3.20 )+- (0.0, 0.3)
( 10, 3.82 )+- (0.0, 0.3)
( 11, 4. 28 )+- (0.0, 0.3)
};  \addlegendentry{Case  1} ;
\addplot+[error bars/.cd,
y dir=both,y explicit]
 coordinates {
( 4 ,0.79 )+- (0.0, 0.6)
( 5, 1.23 )+- (0.0, 0.5)
( 6, 1.92 )+- (0.0, 0.3)
( 7, 2.35 )+- (0.0, 0.4)
( 8, 3.0 )+- (0.0, 0.3)
( 9, 3.71 )+- (0.0, 0.3)
( 10, 4.0 )+- (0.0, 0.3)
( 11, 4.40 )+- (0.0, 0.3)
};  \addlegendentry{Case 2} ;
\end{axis}
\end{tikzpicture}
\end{minipage}\hfill
\caption{\textcolor{black}{Wall-clock times of the experiments described in Section 5. The simulation settings are the same as before. Here, we only consider random-initialization based experiments. Case 1 and 2 corresponds to the two different hypergraphons considered and dense and sparse refers to the values of $\rho_n$ set in the experiments. }}
\label{fig:wallclock}
\end{figure*}
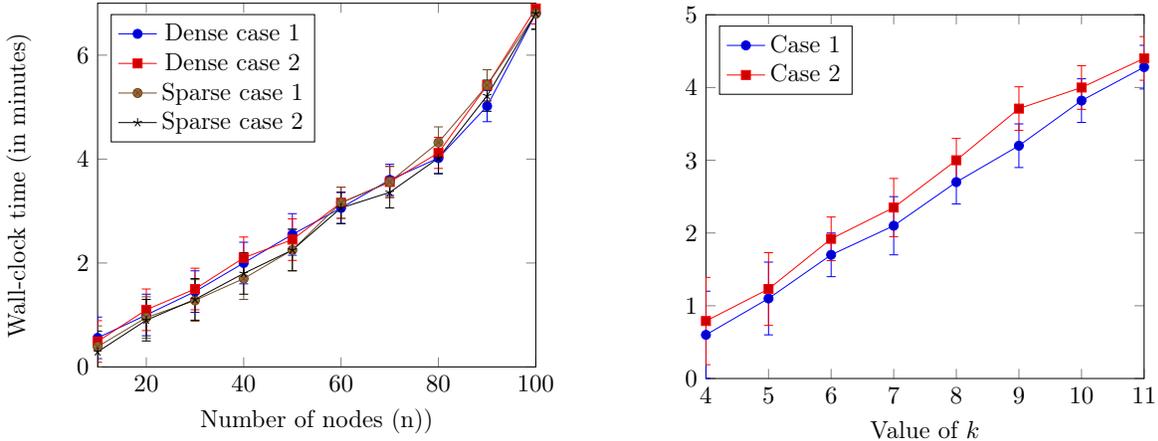

\textcolor{black}{All the simulation settings are as described in Section~\ref{sec:simexp}. Figure~\ref{fig:wallclock} (left) shows the wall-clock time of our proposed algorithm as a function of number of nodes for a fixed value of $k$, corresponding to the hypergraphon in case 1 and case 2 in both the dense and sparse settings. Similarly, Figure~\ref{fig:wallclock} (right) shows the wall-clock time of our proposed algorithm as a function of $k$ for a fixed value of $n$ for the dense setting. We notice that for a fixed $k$, as $n$ is increased, the wall-clock time increases near-linearly first and then starts to increase rapidly. For a fixed $n$, as $k$ increases the  wall-clock time increases near-linearly. It is intriguing to come up with faster implementations of the algorithm, or even better, faster algorithms for estimating SLH hypergraphons.}

\section{Simulation Results: Misspecified Case}\label{sec:misspec}

In this section we motivate the use of simple hypergraphons by demonstrating that they can be good approximations to general hypergraphons. We consider the most elementary hypergraphons for this experiment -- piecewise constant hypergraphons. First we specify the model through a multistage construction, and then formalize the model by constructing its associated hypergraphon. We keep $m=3$ fixed for this experiment.

Let $n$ be the number of vertices, and $k$ be the number of communities. We have four additional parameters, $(p_1, q_1, p_2, q_2)$. Intuitively speaking, depending on vertex communities, we will construct edges between pairs of vertices, which can be viewed as latent variables, and then construct hyperedges between triples of vertices depending on those latent edges, but not directly on the communities of the constituent nodes. First we assign each vertex to one of the $k$ groups uniformly at random. Then between each pair of vertices, we put an edge with probability $p_1$ if the vertices belong to the same community, and with probability $q_1$ otherwise. Next, between each triple of vertices, we put a hyperedge with probability $p_2$ if all three latent edges are present, and with probability $q_2$ otherwise. See \cite{zhao2015hypergraph} for a hypergraphon construction of a similar flavor.

We now construct a (six-dimensional) hypergraphon which matches the distribution of the above model. We break the function into ten cases which correspond to the more intuitive procedural construction we just considered. We begin by defining the sets $A_1, A_2, \ldots, A_5$ and $B_1, B_2, \ldots, B_5$. To aid in our exposition, let $I_x$ be the $i$ such that $x \in \left[\frac{i-1}{k}, \frac{i}{k}\right)$. $I_x = I_y$ will mean that the vertices corresponding to $x$ and $y$ are in the same community. The set $A_1$ corresponds to the event where all three vertices are in the same community. $A_2$ will correspond to the event where the first two vertices are in the same community, $A_3$ to where the first and third are in the same community, and $A_4$ when the second and third are. $A_5$ corresponds to the event where all of the vertices are in the same community. For every assignment of communities to vertices, we have different probabilities of all three edges being present. When all of the edges are present, we have a probability $p_2$ of the hyperedge being in the graph, a probability of $q_2$ otherwise. $B_i$ will correspond to the event where all three edges are present in case $A_i$. Finally, we define our hypergraphon $f$ as displayed in Figure~\ref{fig:mishypergraphon}.

\begin{figure}
{\Large
\begin{align*}
f(x_1, &x_2, x_3, x_{12}, x_{13}, x_{23}) \\
= &\left\{ \begin{array}{ll}  
		p_2 & \text{for } (x_1, x_2, x_3) \in A_1, (x_{12}, x_{13}, x_{23}) \in B_1 \\
		q_2 & \text{for } (x_1, x_2, x_3) \in A_1, (x_{12}, x_{13}, x_{23}) \in B_1^C \\
		p_2 & \text{for } (x_1, x_2, x_3) \in A_2, (x_{12}, x_{13}, x_{23}) \in B_2 \\
		q_2 & \text{for } (x_1, x_2, x_3) \in A_2, (x_{12}, x_{13}, x_{23}) \in B_2^C \\
		p_2 & \text{for } (x_1, x_2, x_3) \in A_3, (x_{12}, x_{13}, x_{23}) \in B_3 \\
		q_2 & \text{for } (x_1, x_2, x_3) \in A_3, (x_{12}, x_{13}, x_{23}) \in B_3^C \\
		p_2 & \text{for } (x_1, x_2, x_3) \in A_4, (x_{12}, x_{13}, x_{23}) \in B_4 \\
		q_2 & \text{for } (x_1, x_2, x_3) \in A_4, (x_{12}, x_{13}, x_{23}) \in B_4^C \\
		p_2 & \text{for } (x_1, x_2, x_3) \in A_5, (x_{12}, x_{13}, x_{23}) \in B_5 \\
		q_2 & \text{for } (x_1, x_2, x_3) \in A_5, (x_{12}, x_{13}, x_{23}) \in B_5^C    \end{array}\right.
\end{align*}
\begin{align*}
&~~~~~~~~A_1 = \{ (x_1, x_2, x_3) \in [0,1]^3 \text{ s.t. } I_{x_1} = I_{x_2} = I_{x_3} \} \\
&~~~~~~~~A_2 = \{ (x_1, x_2, x_3) \in [0,1]^3 \text{ s.t. } I_{x_1} = I_{x_2} \neq I_{x_3} \} \\
&~~~~~~~~A_3 = \{ (x_1, x_2, x_3) \in [0,1]^3 \text{ s.t. } I_{x_1} = I_{x_3} \neq I_{x_2} \} \\
&~~~~~~~~A_4 = \{ (x_1, x_2, x_3) \in [0,1]^3 \text{ s.t. } I_{x_2} = I_{x_3} \neq I_{x_1} \} \\
&~~~~~~~~A_5 = \{ (x_1, x_2, x_3) \in [0,1]^3 \text{ s.t. } I_{x_1} \neq I_{x_2}, I_{x_1} \neq I_{x_3}, I_{x_2} \neq I_{x_3} \} \\
&~~~~~~~~B_1 = \{ (x_{12}, x_{13}, x_{23}) \in [0,p_1) \times [0,p_1) \times [0,p_1) \} \\
&~~~~~~~~B_2 = \{ (x_{12}, x_{13}, x_{23}) \in [0,p_1) \times [0,q_1) \times [0,q_1) \} \\
&~~~~~~~~B_3 = \{ (x_{12}, x_{13}, x_{23}) \in [0,q_1) \times [0,p_1) \times [0,q_1) \} \\
&~~~~~~~~B_4 = \{ (x_{12}, x_{13}, x_{23}) \in [0,q_1) \times [0,q_1) \times [0,p_1) \} \\
&~~~~~~~~B_5 = \{ (x_{12}, x_{13}, x_{23}) \in [0,q_1) \times [0,q_1) \times [0,q_1) \} 
\end{align*}
}
\caption{The full hypergraphon used for testing model-misspecification error.}
\label{fig:mishypergraphon}
\end{figure}

\begin{figure}
\label{fig:misspecified}
\centering
 \includegraphics[scale=0.5]{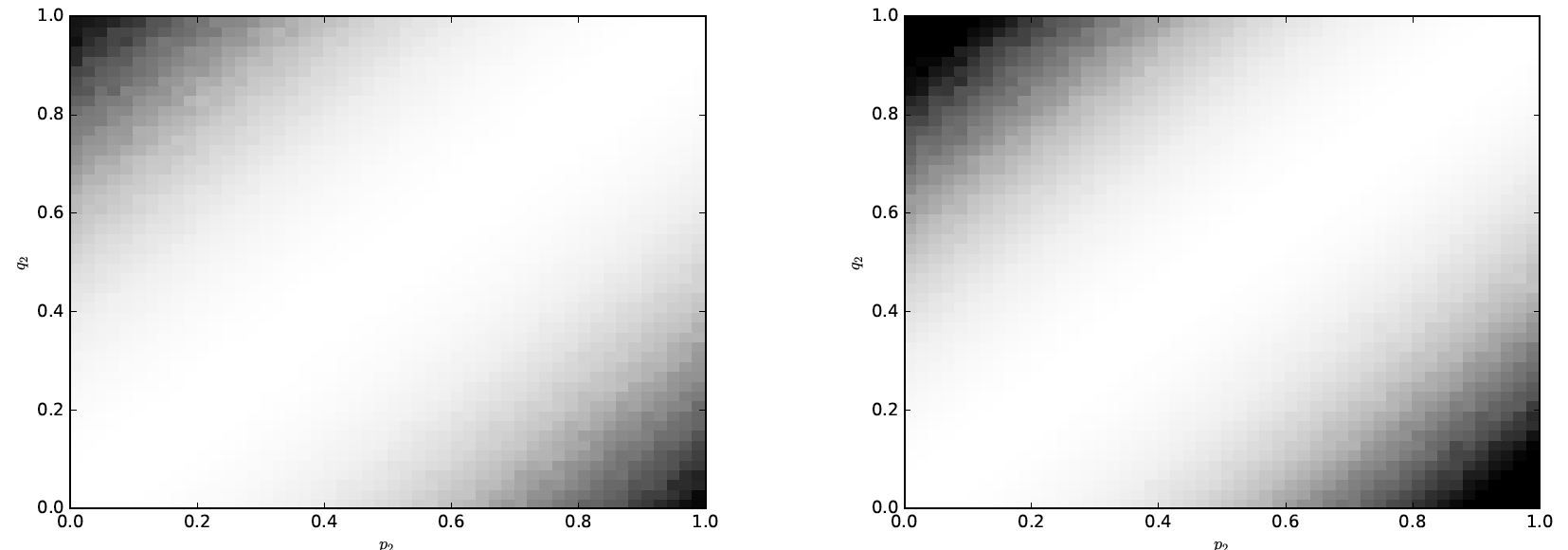}  
\caption{Left figure corresponds to the case of $p_1 = 0.6$ and $q_1= 0.4$ respectively. Right figure corresponds to the case of $p_1 = 0.8$ and $q_1=0.4$. We note that the structure is similar, but as suggested by other figure, when $p_1$ and $q_1$ are big, the error is large. \textcolor{black}{Whiter regions correspond to small normalize  reconstruction error (around 0.2) and darker regions correspond to large estimation errors (around 0.8). }}
\end{figure}

\subsection{Results}
\textcolor{black}{We considered two cases for mis-specified models: case of $p_1 = 0.6$ and $q_1= 0.4$ and the case of $p_1 = 0.8$ and $q_1=0.4$. The number of nodes $n$ was set to be 100. We found that the the choice of $k=0.5\cdot n^3$ gave the best result. We emphasize here that this choice was found through trial and error as there is no supporting theory in the mis-specified case.} Our results for estimation are plotted in Figures 5. We find that in certain regimes of parameters, our estimation algorithm performs well. Firstly, when $p_2$ is close to $q_2$, the graphon resembles a constant hypergraphon and our estimation algorithm performs perfectly. To see this, notice that when $p_2 = q_2$, the resulting hypergraphon is in fact the constant hypergraphon with parameter $p_2 = q_2$. This suggests that our estimation algorithm works when the more complex hypergraphon is structurally close to a simple hypergraphon. We notice that as $p_1, q_1$ get bigger, so does the estimation error. This is because when $p_1$ and $q_1$ are both small, the resulting hypergraphon resembles the constant hypergraphon with parameter $q_2$, since edges are rare. For larger $p_1,q_1$, the resulting hypergraphon is too rich for our estimation procedure to handle, and we observe larger estimation error.

Finally, we observed that estimation succeeds when community detection fails, a common phenomenon since estimation is generally an easier problem than community detection. However, we also observe that community detection often succeeds when estimation fails, when $|p_2 - q_2|$ is large for example. This is telling because our estimation algorithm first does community detection, and estimates $Q$ from the estimated communities. We can conclude that this is in fact model misspecification error, since estimation fails even when community detection succeeds.






\newpage
\section{Applications to Hyperedge Prediction}

In this section, we demonstrate the applicability of the proposed hypergraphon estimator in Sections~\ref{sec:mainresult} and~\ref{sec:algorithm} to the problem of hyperedge prediction (also called hyperlink prdiction) in $m$-uniform hypergraphs. Here, we assume that we are not given the entire adjacency tensor $\mathbf{A}$. Instead, we only observe a fraction of entries; the set of observed entries is denoted as $\Omega$. Based on the observed entries, the task is to predict the presence or absence of the missing hyperedges. In order to do so, we modify the estimator in~\eqref{ref:estimator} as 
$
\hat{\mathbf{\Theta}}= \argmin_{\Theta \in T_z} \{ \|\Theta\|^2_F  - \frac{2n^m}{|\Omega|} \sum_{j \in \Omega} \mathbf{A}_j \mathbf{\Theta}_j\},
$
in order to account for missing entries in the adjacency tensor. A similar method was also suggested by~\cite{gao2015rate} for the graph setting. Based on the estimated $\hat{\mathbf{\Theta}}$, we predict the missing hyperedge as present if the corresponding entry in the tensor $\hat{\mathbf{\Theta}}$ is greater than 0.5 and absent otherwise.

\begin{table}
\centering
\begin{tabular}{|l|lll|lll|}
    \hline
    \multirow{2}{*}{} &
      \multicolumn{3}{c}{\texttt{GPS}} &
      \multicolumn{3}{c|}{\texttt{MovieLens}}\\
   & $70\%$ & $80\%$ & $90\%$ & $70\%$ & $80\%$ & $90\%$ \\
    \hline
    CMM& $0.64 \pm0.03$ &  $0.70 \pm 0.05$&  $0.79 \pm0.05$ & $0.76 \pm 0.06$& $0.82 \pm 0.05  $ & $0.88 \pm 0.06$ \\
    \hline
 C3MM    & $0.69 \pm 0.04$ & $0.76 \pm 0.03$ & $0.83 \pm 0.05$ &  $0.81 \pm 0.06$ & $0.87 \pm 0.06$  & $0.90 \pm 0.04$ \\
    \hline
    Hypergraphon &$ 0.73 \pm 0.03$ & $0.76 \pm 0.05$ & $0.82 \pm 0.02 $& $0.85 \pm 0.06$& $0.89 \pm 0.04$ & $0.90 \pm 0.06$ \\
    \hline
\end{tabular}
\caption{Area under ROC curve for hyperedge prediction on \texttt{GPS} and \texttt{MovieLens} datasets.}
\label{tab:edgepred}
\end{table}

For our experiments, we use the the \texttt{GPS} dataset~\citep{zheng2010collaborative} and the \texttt{MovieLens} dataset~\citep{harper2015movielens}. The \texttt{GPS} dataset consists of 146 users, at 70 locations, performing 5 different types of activities. This dataset consists of 1436 hyperedges in total that were $1$s. For the \texttt{MovieLens} data, we used a smaller version with 500 users, 500 movies and 100 tags, for computational simplicity. This dataset consists of 8,349 hyperedges that were $1$s. For the sake of experiments, we randomly pick $70\%$, $80\%$ and $90\%$ and treat them as observed data. The problem is to predict the missing hyperedges. The value of $k$ for the hypergaphon method was set as the one that obtains the best prediction performance, by trial-and-error method. We remark that developing principled approaches (e.g., cross-validation) for hyperparameter selection with network data is an interesting problem (which has generated great interest in the community recently), which, however is beyond the scope of this work. 
We compare the performance of our method against two factorization-based methods termed as CMM~\citep{zhang2018beyond} and C3MM~\citep{sharmac3mm}. The above methods were chosen as they are recent works for hyperedge prediction with superior or comparable performance to several baselines. We performed the experiment for 50 trials and calculated the average Area under ROC curve (AUC) as a measure of performance. From the results in Table~\ref{tab:edgepred}, we see that the hypergraphon based approach is comparable to the CMM and C3MM method with demonstrating superiority when the percentage of missing hyperedges in the given hypergraph is large.


\section{Discussion}
In this paper, we initiated the study of hypergraphon for nonparametric modeling of hypergraphs. We provided rates of convergence in expectation for estimating a class of  Smooth Lipschitz Hypergraphons (SLH) and provided practical algorithms for implementing the estimators. There are several directions for future work, some of which we highlight below.\\

\noindent \textbf{Estimating the graphon function}: Note that in this paper, we consider estimating the probability of formation of hyperedges in the $L_2$ norm. A more challenging problem is to estimate the hypergraphon function itself from the given adjacency tensor. In order to do so, one needs to define appropriate norms on the space of hypergraphons. For the case of graphons, the cut-norm, defined as $$ \|f\|_\square =  \underset{S,T \subseteq [0,1]}{\sup} \left| \int_{S \times T} f(x,y) \, dx\, dy\right|,$$
where $S$ and $T$ are measurable subset of $[0,1]$ serves as a good metric; see~\cite{klopp2017optimal}. But in the case of hypergraphons the problem is more delicate. A straight forward generalization of the above cut norms (defined below for simple $3$-uniform hypergraphons)
 $$ \|f\|_\square =  \underset{S,T,U \subseteq [0,1]}{\sup} \left| \int_{S \times T \times U} f(x,y,z) \, dx\, dy \, dz\right|,$$
 provides only a weaker metric on the space of simple hypergraphons. Care must be taken first to define first a metric that is meaningful; see~\cite{zhao2015hypergraph}. We leave this problem of providing cut-norm convergence results for estimating hypergraphons as future work.\\
 
\noindent \textbf{Computational Theory}: The algorithm presented in section~\ref{sec:algorithm} works well as shown via the experimental results, surprisingly also with random initialization in spite of being a non-convex optimization problem. Recent advances in non-convex optimization, also highlights a similar phenomenon holds for other models; see, for example,~\cite{Chen2019} for the case of phase retrieval model. It is interesting to leverage such results in the context of hypergraphon models. Furthermore, for the case of graphons,~\cite{zhang2017estimating} proposed a neighborhood smoothing approach in particular for estimating the probability matrix. It is not clear if such an approach is immediately extendable for hypergraphons. But it is interesting to explore if a similar approach could be leveraged for hypergraphons. \\

\noindent  \textbf{2-layer interaction model}: A direct step-function (or SBM) based approximation of the full hypergraphon is provided in Equation 13 in~\cite{zhao2015hypergraph}. It would be interesting to explore to use a layer wise approach based on the Q-step approximation to estimate the full hypergraphon function in the future. \\

\noindent \textbf{Hierarchical clustering}: Based on the idea of graphons, recently a hierarchical clustering framework based was proposed in~\cite{eldridge2016graphons}. Also model-free interpretations of the standard clustering algorithm was provided in~\cite{diao2016model}. It would extremely interesting to explore similar extension to the case of hypergraphons. 

\appendix{}
\section{Proofs for Section~\ref{sec:mainresult}}\label{sec:supp}

In this section, we provide the proofs for Lemma~\ref{lem:est} and Lemma~\ref{lem:approx} from Section~\ref{sec:mainresult}. Throughout the proof we assume that $C$ is a absolute constant that changes from step to step. Furthermore, in this section, we denote $\E_{\mathbf{A},X}$ and $\E_{X}$, as just $\E$ for simplicity, as it will be clear from the context. \textcolor{black}{We also require some additional notations: Recall that $G \subset [n]^m$ denotes the set of hyperedges with no repeated vertices. That is, $j \not\in G$ iff $\exists~ u,v$ s.t $j_u = j_v$. We define $H \in [n]^m$ to be the set of increasing hyperedges. That is, $j \in H$ iff $j_1 < j_2 < \ldots < j_m$. Then, note that $|H| = \binom{n}{m}$ and $|G| = m!|H|$.}

\begin{proof}[Proof of Lemma~\ref{lem:est}] First, we consider the loss function
\begin{align*}
	L(\mathbf{Q},z) = \frac12 \sum_{a \in [k]^{m}} \sum_{j \in z^{-1}(a) \cap H} (\mathbf{A}_{j} - \mathbf{Q}_{a})^2
\end{align*}
Let $\hat{\mathbf{Q}}$, $\hat z$ be minimizers of $L(\mathbf{Q},z)$ and let $\hat{\mathbf{\Theta}}$ be such that $\hat{\mathbf{\Theta}}_{j} = \hat{\mathbf{Q}}_{\hat z(j)}$. Note that this is a least squares estimator. Let $\mathbf{\Theta}_*$ be the best $k$-class block model approximation of $\mathbf{\bar{\Theta}} $ in Frobenius norm. In the case where $\mathbf{\bar{\Theta}} $ is truly a $k$-class block model, $\mathbf{\Theta}_* = \mathbf{\bar{\Theta}} $. Since $\hat{\mathbf{\Theta}}$ is a least squares estimator, we have the following set of inequalities:
\begin{align*}
			     & \| \hat{\mathbf{\Theta}} - \mathbf{A} \|_F^2  \leq \| \mathbf{\Theta}_* - \mathbf{A} \|_F^2 \\
	\Leftrightarrow & \| \hat{\mathbf{\Theta}} - \mathbf{\bar{\Theta}}  \|_F^2 + 2\langle \hat{\mathbf{\Theta}} - \mathbf{\bar{\Theta}} , \mathbf{\bar{\Theta}}  - \mathbf{A} \rangle + \| \mathbf{\bar{\Theta}}  - \mathbf{A} \|_F^2 \leq \| \mathbf{\Theta}_* - \mathbf{\bar{\Theta}}  \|_F^2 + 2\langle \mathbf{\Theta}_* - \mathbf{\bar{\Theta}} , \mathbf{\bar{\Theta}}  - \mathbf{A} \rangle + \| \mathbf{\bar{\Theta}}  - \mathbf{A} \|_F^2 \\
	\Leftrightarrow &\| \hat{\mathbf{\Theta}} - \mathbf{\bar{\Theta}}  \|_F^2   \leq \| \mathbf{\Theta}_* - \mathbf{\bar{\Theta}}  \|_F^2 + 2\langle \hat{\mathbf{\Theta}} - \mathbf{\bar{\Theta}} , \mathbf{A} - \mathbf{\bar{\Theta}}  \rangle + 2\langle \mathbf{\bar{\Theta}}  - \mathbf{\Theta}_*, \mathbf{A} - \mathbf{\bar{\Theta}}  \rangle \\
	\Leftrightarrow& \| \hat{\mathbf{\Theta}} - \mathbf{\bar{\Theta}}  \|_F^2 \leq \| \mathbf{\Theta}_* - \mathbf{\bar{\Theta}}  \|_F^2 + 2\langle \hat{\mathbf{\Theta}} - \mathbf{\bar{\Theta}} , \mathbf{E} \rangle + 2\langle \mathbf{\bar{\Theta}}  - \mathbf{\Theta}_*, \mathbf{E} \rangle,
\end{align*}
where $\mathbf{E} = \mathbf{A} - \mathbf{\bar{\Theta}}$ is the noise tensor. Now we are interesting in bounding the expectation of the left hand side. Since $\mathbf{\Theta}_*$ and $\mathbf{\bar{\Theta}} $ are deterministic and $\E[\mathbf{E}] = 0$, the final summand has zero mean, so it suffices to bound the expectation of $\langle \hat{\mathbf{\Theta}} - \mathbf{\bar{\Theta}} , \mathbf{E} \rangle$. For any $z \in \mathcal Z_{n.k}$, let $\tilde{\mathbf{\Theta}}_z$ be the best Frobenius norm approximation of $\mathbf{\bar{\Theta}} $ in the collection of rank $m$ tensors
\[
	\mathcal T_z = \{ \mathbf{\Theta} : \exists \mathbf{Q} \in \R_{\text{sym}}^{k \times \cdots \times k} \text{ such that } \mathbf{\Theta}_j = Q_{z(j)} \text{ for } j \in G, \text{ and } \mathbf{\Theta}_j = 0 \text{ otherwise} \}
\]

Note that we can interpret $\mathcal T_z$ as the collection of $\mathbf{\Theta}$'s which correspond to a $k$-class block model with classes given by $z$. More concretely, $\tilde{\mathbf{\Theta}}_z$ is constructed by taking blockwise averages of $\mathbf{\bar{\Theta}} $ according to $z$, the same way $\hat{\mathbf{\Theta}}_z$ is constructed by taking blockwise averages of $\mathbf{A}$ with respect to $z$. With this interpretation, we have the decomposition 
\[
	\langle \hat{\mathbf{\Theta}} - \mathbf{\bar{\Theta}} , \mathbf{E} \rangle = \langle \tilde{\mathbf{\Theta}}_z - \mathbf{\bar{\Theta}} , \mathbf{E} \rangle + \langle \hat{\mathbf{\Theta}} - \tilde{\mathbf{\Theta}}_z, \mathbf{E} \rangle.
\]
Here the first summand, $\langle \tilde{\mathbf{\Theta}}_z - \mathbf{\bar{\Theta}} , \mathbf{E} \rangle$ corresponds to the error incurred from misclustering, and $\langle \hat{\mathbf{\Theta}} - \tilde{\mathbf{\Theta}}_z, \mathbf{E} \rangle$ corresponds to error from Bernoulli noise. We bound each of the two terms separately next.

\noindent \textbf{Control of $\langle \tilde{\mathbf{\Theta}}_z - \mathbf{\bar{\Theta}} , \mathbf{E} \rangle$} : We first express $\langle \tilde{\mathbf{\Theta}}_{\hat z} - \mathbf{\bar{\Theta}} , \mathbf{E} \rangle$ as a sum of independent random variables. To do this we only sum over $j$ which are in increasing order, and multiply by the appropriate factor $m!$. Hence we have
\[
	\langle \tilde{\mathbf{\Theta}}_{\hat z} - \mathbf{\bar{\Theta}} , \mathbf{E} \rangle = \sum_{j \in [n]^m}(\tilde{\mathbf{\Theta}}_{\hat z} - \mathbf{\bar{\Theta}} )_j \mathbf{E}_j = \sum_{j \in H} m! (\tilde{\mathbf{\Theta}}_{\hat z} - \mathbf{\bar{\Theta}} )_j \mathbf{E}_j.
\]
Note $|H| = \binom{n}{m}$, and that the summands are only positive in $G$. We use the above decomposition, and Bernstein's inequality (see Theorem~\ref{thm:bern} below), for a fixed, deterministic $z$ to get following tail bounds:
\begin{align*}
	&\P\left(\langle \tilde{\mathbf{\Theta}}_z - \mathbf{\bar{\Theta}} , \mathbf{E} \rangle \geq \sqrt{2t \sum_{j \in H} \Var(m!(\tilde{\mathbf{\Theta}}_z - \mathbf{\bar{\Theta}} )_j\mathbf{E}_j)} + \frac{2m!\| \tilde{\mathbf{\Theta}}_z - \mathbf{\bar{\Theta}}  \|_{\infty}}{3} \right) \leq e^{-t} \\
	\Leftrightarrow&\P\left(\langle \tilde{\mathbf{\Theta}}_z - \mathbf{\bar{\Theta}} , \mathbf{E} \rangle \geq \sqrt{2t (m!)^2 \sum_{j \in H} (\tilde{\mathbf{\Theta}}_z - \mathbf{\bar{\Theta}} )_j^2 \Var(\mathbf{E}_j)} + \frac{2m!\| \tilde{\mathbf{\Theta}}_z - \mathbf{\bar{\Theta}}  \|_{\infty}}{3} \right) \leq e^{-t}.
\end{align*}
Now note that the variance in the above expression is bound as $$\Var(\mathbf{E}_j) = (\mathbf{\bar{\Theta}} )_j(1-(\mathbf{\bar{\Theta}} )_j) \leq (\mathbf{\bar{\Theta}} )_j \leq \| \mathbf{\bar{\Theta}}  \|_\infty.$$ 
Hence substituting this expression, we have 
\begin{align*}
	&\P\left(\langle \tilde{\mathbf{\Theta}}_z - \mathbf{\bar{\Theta}} , \mathbf{E} \rangle \geq \sqrt{2t (m!)^2 \sum_{j \in H} (\tilde{\mathbf{\Theta}}_z - \mathbf{\bar{\Theta}} )_j^2 \| \mathbf{\bar{\Theta}}  \|_\infty} + \frac{2m!\| \tilde{\mathbf{\Theta}}_z - \mathbf{\bar{\Theta}}  \|_{\infty}}{3} \right) \leq e^{-t} \\
	\Leftrightarrow&\P\left(\langle \tilde{\mathbf{\Theta}}_z - \mathbf{\bar{\Theta}} , \mathbf{E} \rangle \geq \sqrt{2t  \| \mathbf{\bar{\Theta}}  \|_\infty (m!)^2 \sum_{j \in H} (\tilde{\mathbf{\Theta}}_z - \mathbf{\bar{\Theta}} )_j^2 } + \frac{2m!\| \tilde{\mathbf{\Theta}}_z - \mathbf{\bar{\Theta}}  \|_{\infty}}{3} \right) \leq e^{-t} \\
	\Leftrightarrow&\P\left(\langle \tilde{\mathbf{\Theta}}_z - \mathbf{\bar{\Theta}} , \mathbf{E} \rangle \geq \sqrt{2t  \| \mathbf{\bar{\Theta}}  \|_\infty m! \| \tilde{\mathbf{\Theta}}_z - \mathbf{\bar{\Theta}}  \|_F^2 } + \frac{2m!\| \tilde{\mathbf{\Theta}}_z - \mathbf{\bar{\Theta}}  \|_{\infty}}{3} \right) \leq e^{-t} \\
	\Leftrightarrow&\P\left(\langle \tilde{\mathbf{\Theta}}_z - \mathbf{\bar{\Theta}} , \mathbf{E} \rangle \geq \| \tilde{\mathbf{\Theta}}_z - \mathbf{\bar{\Theta}}  \|_F \sqrt{2m!  \| \mathbf{\bar{\Theta}}  \|_\infty t } + \frac{2m!\| \tilde{\mathbf{\Theta}}_z - \mathbf{\bar{\Theta}}  \|_{\infty}}{3} \right) \leq e^{-t} .
\end{align*}
Now we use a union bound to get a uniform in $z$ version of the above expression tail bound. To do so, let 
\[
	A_z = \left\{ \langle \tilde{\mathbf{\Theta}}_z - \mathbf{\bar{\Theta}} , \mathbf{E} \rangle \geq \| \tilde{\mathbf{\Theta}}_z - \mathbf{\bar{\Theta}}  \|_F \sqrt{2m!  \| \mathbf{\bar{\Theta}}  \|_\infty t } + \frac{2m!\| \tilde{\mathbf{\Theta}}_z - \mathbf{\bar{\Theta}}  \|_{\infty}}{3} \right\},
\]
and notice that $A_{\hat z} \subset  \underset{z \in \mathcal Z_{n,k}}{\bigcup} A_z$. We proceed with a union bound to get the following bound. 
\[
	\P(A_{\hat z}) \leq \P\left( \bigcup_{z \in \mathcal Z_{n,k}} A_z \right) \leq \sum_{z \in \mathcal Z_{n,k}} \P(A_z) \leq k^n e^{-t} = e^{-(t - n\log k)}.
\]
Substituting $\tilde t = t - n\log k$ and then relabeling $\hat t$ by $t$ yields the following:
\begin{align*}
	&\P\left(\langle \tilde{\mathbf{\Theta}}_{\hat z} - \mathbf{\bar{\Theta}} , \mathbf{E} \rangle \geq \| \tilde{\mathbf{\Theta}}_{\hat z} - \mathbf{\bar{\Theta}}  \|_F \sqrt{2m!  \| \mathbf{\bar{\Theta}}  \|_\infty t } + \frac{2m!\| \tilde{\mathbf{\Theta}}_{\hat z} - \mathbf{\bar{\Theta}}  \|_{\infty}}{3} \right) \leq e^{-(t - n\log k)} \\
	\Leftrightarrow&\P\left(\langle \tilde{\mathbf{\Theta}}_{\hat z} - \mathbf{\bar{\Theta}} , \mathbf{E} \rangle \geq \| \tilde{\mathbf{\Theta}}_{\hat z} - \mathbf{\bar{\Theta}}  \|_F \sqrt{2m!  \| \mathbf{\bar{\Theta}}  \|_\infty (\tilde t + n\log k) } + \frac{2m!\| \tilde{\mathbf{\Theta}}_{\hat z} - \mathbf{\bar{\Theta}}  \|_{\infty}}{3} \right) \leq e^{-\tilde t} \\
	\Leftrightarrow&\P\left(\langle \tilde{\mathbf{\Theta}}_{\hat z} - \mathbf{\bar{\Theta}} , \mathbf{E} \rangle \geq \| \tilde{\mathbf{\Theta}}_{\hat z} - \mathbf{\bar{\Theta}}  \|_F \sqrt{2m!  \| \mathbf{\bar{\Theta}}  \|_\infty (t + n\log k) } + \frac{2m!\| \tilde{\mathbf{\Theta}}_{\hat z} - \mathbf{\bar{\Theta}}  \|_{\infty}}{3} \right) \leq e^{-t} .
\end{align*}
Since $\tilde{\mathbf{\Theta}}_{\hat z}$ is an averaging of $\mathbf{\Theta}$ over blocks, we have $\| \tilde{\mathbf{\Theta}}_{\hat z} - \mathbf{\bar{\Theta}}  \|_{\infty}$. Furthermore, using the inequality $2uv \leq u^2 + v^2$, we have $$\| \tilde{\mathbf{\Theta}}_{\hat z} - \mathbf{\bar{\Theta}}  \|_F \sqrt{2m!  \| \mathbf{\bar{\Theta}}  \|_\infty (t + n\log k) } \leq \frac18\| \tilde{\mathbf{\Theta}}_{\hat z} - \mathbf{\bar{\Theta}}  \|_F^2 + 4m!\| \mathbf{\bar{\Theta}}  \|_\infty (t + n\log k). $$Combining these two observations yields
\begin{align*}
	&\P\left(\langle \tilde{\mathbf{\Theta}}_{\hat z} - \mathbf{\bar{\Theta}} , \mathbf{E} \rangle \geq \frac18\| \tilde{\mathbf{\Theta}}_{\hat z} - \mathbf{\bar{\Theta}}  \|_F^2 + 4m!\| \mathbf{\bar{\Theta}}  \|_\infty (t + n\log k) + \frac{2m!\| \mathbf{\bar{\Theta}}  \|_{\infty}}{3} \right) \leq e^{-t} \\
	\Leftrightarrow&\P\left(\langle \tilde{\mathbf{\Theta}}_{\hat z} - \mathbf{\bar{\Theta}} , \mathbf{E} \rangle - \frac18\| \tilde{\mathbf{\Theta}}_{\hat z} - \mathbf{\bar{\Theta}}  \|_F^2 \geq 4m!\| \mathbf{\bar{\Theta}}  \|_\infty (t + n\log k) + \frac{2m!\| \mathbf{\bar{\Theta}}  \|_{\infty}}{3} \right) \leq e^{-t}.
\end{align*}
Finally letting $u = 4m! \| \mathbf{\bar{\Theta}}  \|_\infty (t + n\log k) + \frac{2m!\| \mathbf{\bar{\Theta}}  \|_\infty}{3}$, we obtain 
\[
	\P\left(\langle \tilde{\mathbf{\Theta}}_{\hat z} - \mathbf{\bar{\Theta}} , \mathbf{E} \rangle - \frac18\| \tilde{\mathbf{\Theta}}_{\hat z} - \mathbf{\bar{\Theta}}  \|_F^2 \geq u \right) \leq e^{n\log k + 1/6 - \frac{3u}{4m!\| \mathbf{\bar{\Theta}}  \|_\infty}}.
\]
Now we proceed to obtain the following expectation bound from the above probability bound. Integrating both sides with respect to $u$ gives us 
\begin{align*}
	\E[\langle \tilde{\mathbf{\Theta}}_{\hat z} - \mathbf{\bar{\Theta}} , \mathbf{E} \rangle - \frac18\| \tilde{\mathbf{\Theta}}_{\hat z} - \mathbf{\bar{\Theta}}  \|_F^2] &\leq \int_0^{u^*} \P\left(\langle \tilde{\mathbf{\Theta}}_{\hat z} - \mathbf{\bar{\Theta}} , \mathbf{E} \rangle - \frac18\| \tilde{\mathbf{\Theta}}_{\hat z} - \mathbf{\bar{\Theta}}  \|_F^2 \geq u \right) du \\
	&~~~+ \int_{u^*}^\infty \P\left(\langle \tilde{\mathbf{\Theta}}_{\hat z} - \mathbf{\bar{\Theta}} , \mathbf{E} \rangle - \frac18\| \tilde{\mathbf{\Theta}}_{\hat z} - \mathbf{\bar{\Theta}}  \|_F^2 \geq u \right) du \\
					&\leq u^* + \int_{u^*}^\infty e^{n\log k + 1/6 - \frac{3u}{4m!\| \mathbf{\bar{\Theta}}  \|_\infty}} du \\
					&= u^* + \left.\frac{m!\| \mathbf{\bar{\Theta}}  \|_\infty}{2} e^{n\log k + 1/6 - \frac{3u}{4m!\| \mathbf{\bar{\Theta}}  \|_\infty}}\right|_\infty^{u^*} \\
					&= u^* + \frac{m!\| \mathbf{\bar{\Theta}}  \|_\infty}{2} e^{n\log k + 1/6 - \frac{3u^*}{4m!\| \mathbf{\bar{\Theta}}  \|_\infty}}
\end{align*}
Choosing $u^* = Cm!\| \mathbf{\bar{\Theta}}  \|_\infty n\log k$ yields the following:
\begin{align}\label{eq:control1}
	\E[\langle \tilde{\mathbf{\Theta}}_{\hat z} - \mathbf{\bar{\Theta}} , \mathbf{E} \rangle - \frac18\| \tilde{\mathbf{\Theta}}_{\hat z} - \mathbf{\bar{\Theta}}  \|_F^2] \leq Cm!\| \mathbf{\bar{\Theta}}  \|_\infty n\log k
\end{align}
which completes the control of $\langle \tilde{\mathbf{\Theta}}_z - \mathbf{\bar{\Theta}} , \mathbf{E} \rangle$. We now proceed to control the second term.

\textbf{Control of $\langle \hat{\mathbf{\Theta}}_z - \tilde{\mathbf{\Theta}}_z, \mathbf{E} \rangle$} : Controlling this term is more involved than the first term. We give a brief overview of the steps need to get the required bound. We first construct a $1/4$ net to ensure closeness in Frobenius norm, and then modify the net slightly to also ensure closeness in infinity norm. After that we follow the standard Bernstein pipeline as in the control of the first term. Then we handle a more involved bound on $\| \tilde{\mathbf{A}}_{\hat z} - \tilde{\mathbf{\Theta}}_{\hat z} \|_\infty$. We elaborate on the steps below.

For any $z \in \mathcal Z_{n,k}$, define $\mathcal T_{z,1} = \{ \mathbf{\Theta} \in \mathcal T_z : \| \mathbf{\bar{\Theta}}  \|_F \leq 1 \}$ to be the set of all tensors whose block structure is determined by $z$, and whose Frobenius norm is bounded by 1. Denote by $\tilde{\mathbf{A}}_z$ the best Frobenius norm approximation of $\mathbf{A}$ in $\mathcal T_z$, i.e. the blockwise average of $\mathbf{A}$ according to $z$. Then $\tilde{\mathbf{E}}_z = \tilde{\mathbf{A}}_z - \tilde{\mathbf{\Theta}}_z$ is also the projection and blockwise average of $\mathbf{E}$ onto/of $\mathcal T_z$.  By properties of projections, $\mathbf{\Theta} = \frac{\tilde{\mathbf{E}}_z}{\| \tilde{\mathbf{E}}_z \|_F}$ maximizes $\langle \mathbf{\Theta}, \mathbf{E} \rangle$ over $\mathcal T_{z,1}$. Let $C_z$ be a minimal $1/4$ net of $\mathcal T_{z,1}$ in Frobenius norm. Now to each $\mathbf{V} \in C_z$ associate
\[
	\tilde{\mathbf{V}} = \argmin_{\mathbf{\Theta} \in \mathcal T_{z,1} \cap B(\mathbf{V},1/4)} \| \mathbf{\Theta} \|_\infty.
\]
Define $\tilde C_z = \{ \tilde{\mathbf{V}} : \mathbf{V} \in C_z \}$. A standard bound on covering numbers implies $\log |\tilde C_z| \leq Ck^m$. Now using Bernstein inequality (see Theorem~\ref{thm:bern}) and a union bound over $z \in \mathcal Z_{n,k}$ and $\mathbf{\Theta} \in C_z$, we have,
\[
	\P\left( \langle \mathbf{\Theta}, \mathbf{E} \rangle \geq \sqrt{2m! \| \mathbf{\bar{\Theta}}  \|_\infty (t + n\log k + k^m)} + \frac{2m!\| \mathbf{\Theta} \|_\infty}{3} (t + n\log k + k^m)  \right) \leq e^{-t},
\]
where we used the fact that $\| \mathbf{\Theta} \|_F \leq 1$ on $\tilde C_z$. By definition of $\tilde C_z$, there is a $\mathbf{\Theta} \in \tilde C_z$ such that
\[
	\left\| \mathbf{\Theta} - \frac{\tilde{\mathbf{E}}_{\hat z}}{\| \tilde{\mathbf{E}}_{\hat z} \|_F} \right\|_F \leq \frac12 \tab \text{and} \tab \| \mathbf{\Theta} \|_\infty \leq \frac{\| \tilde{\mathbf{E}}_{\hat z} \|_\infty}{\| \tilde{\mathbf{E}}_{\hat z} \|_F} .
\]
Furthermore, for this $\mathbf{\Theta}$, we have $2 \left( \frac{\tilde{\mathbf{E}}_{\hat z}}{\| \tilde{\mathbf{E}}_{\hat z} \|_F} - \mathbf{\Theta} \right)$ belongs to $\mathcal T_{z,1}$. Thus, we have
\[
\left\langle 2 \left( \frac{\tilde{\mathbf{E}}_{\hat z}}{\| \tilde{\mathbf{E}}_{\hat z} \|_F} - \mathbf{\Theta} \right), \mathbf{E}\right\rangle \leq \left\langle \frac{\tilde{\mathbf{E}}_{\hat z}}{\| \tilde{\mathbf{E}}_{\hat z} \|_F}, \mathbf{E}\right\rangle,
\]
or equivalently,
\[
\left\langle \frac{\tilde{\mathbf{E}}_{\hat z}}{\| \tilde{\mathbf{E}}_{\hat z} \|_F}, \mathbf{E} \right\rangle \leq 2\langle \mathbf{\Theta}, \mathbf{E} \rangle.
\]
This gives us
\[
	\langle \tilde{\mathbf{E}}_{\hat z}, \mathbf{E} \rangle \leq 2 \| \tilde{\mathbf{E}}_{\hat z} \|_F \sqrt{2m! \| \mathbf{\bar{\Theta}}  \|_\infty (t + n\log k + k^m)} + \frac{4m!}{3} \| \tilde{\mathbf{E}}_{\hat z} \|_\infty (t + n\log k + k^m)
\]
with probability at least $1 - e^{-t}$. Now we need a bound on $\| \tilde{\mathbf{E}}_{\hat z} \|_\infty$. Every entry of $\tilde{\mathbf{E}}_{\hat z}$ is an average over some block of $\mathbf{E}$, so we take a union bound over all blocks of all legal sizes. First we write what an entry of $\tilde{\mathbf{E}}_{\hat z}$ looks like:
\[
	\left[\tilde{\mathbf{E}}_{\hat z}\right]_j = \frac{\sum _{i \in G : \hat z(i) = \hat z(j)} \mathbf{E}_i  }{ | \{i \in G : \hat z(i) = \hat z(j)\} | }.
\]
Note that we then have the following to be true 
\[
	\| \tilde{\mathbf{E}}_{\hat z} \|_\infty \leq \sup_{s \in S} X_s~\quad\text{and}\quad X_s = \sup_{V \subset [n]^m : |V_i| = s_i} \frac{\sum _{j \in G \cap V } \mathbf{E}_j}{ |G \cap V| },
\]
where $S = \{ n_0, \ldots, n \}^m$.


First we bound $\frac{1}{|G \cap V|} \sum \textbf{E}_j$ using Bernstein's inequality (see Theorem~\ref{thm:bern}), and the inequality $2ab \leq a^2 + b^2$. We need to be careful that we only consider independent summands. We denote by $|\#j|$ the number of permutations of $j$ which are contained in $V$, and use the trivial bound $|\#j| \leq m!$. Based on the above observations, we have the following set of inequalities:
\begin{align*}
	&\P\left( \sum_{j \in G \cap V} \mathbf{E}_j \geq \sqrt{2t \sum_{j \in H \cap V} \Var(|\#j|X_j) } + \frac{2t}{3} \right) \leq e^{-t} \\
	\Rightarrow&\P\left( \sum_{j \in G \cap V} \mathbf{E}_j \geq \sqrt{2tm! \sum_{j \in H \cap V} |\#j| \Var(X_j) } + \frac{2t}{3} \right) \leq e^{-t} \\
	\Rightarrow&\P\left( \sum_{j \in G \cap V} \mathbf{E}_j \geq \sqrt{2tm! \sum_{j \in H \cap V} |\#j| \| \mathbf{\bar{\Theta}}  \|_\infty } + \frac{2t}{3} \right) \leq e^{-t} \\
	\Leftrightarrow&\P\left( \sum_{j \in G \cap V} \mathbf{E}_j \geq \sqrt{2tm! |G \cap V| \| \mathbf{\bar{\Theta}}  \|_\infty } + \frac{2t}{3} \right) \leq e^{-t} \\
	\Leftrightarrow&\P\left( \frac{1}{|G \cap V|} \sum_{j \in G \cap V} \mathbf{E}_j \geq \sqrt{\frac{2tm!}{|G \cap V|} \| \mathbf{\bar{\Theta}}  \|_\infty } + \frac{2t}{3|G \cap V|} \right) \leq e^{-t} \\
	\Rightarrow&\P\left( \frac{1}{|G \cap V|} \sum_{j \in G \cap V} \mathbf{E}_j \geq \| \mathbf{\bar{\Theta}}  \|_\infty + \frac{tm!}{2|G \cap V|}  + \frac{2t}{3|G \cap V|} \right) \leq e^{-t} \\
	\Rightarrow&\P\left( \frac{1}{|G \cap V|} \sum_{j \in G \cap V} \mathbf{E}_j \geq \| \mathbf{\bar{\Theta}}  \|_\infty + \frac{tm!}{|G \cap V|} \right) \leq e^{-t} \\
\end{align*}
By Stirling's formula, we have $|\{ V_i : |V_i| = s_i \}| \leq \binom{n}{s_i} \leq \left(\frac{en}{s_i}\right)^{s_i}$. Therefore, to bound $X_s$, we use a union bound over possible $V_i$, a set which has cardinality at most $\prod_{i=1}^m \left(\frac{en}{s_i}\right)^{s_i}$, which gives us the following:

\[
	\P\left( X_s \geq \| \mathbf{\bar{\Theta}}  \|_\infty + m!\frac{t + \sum_{i=1}^m s_i \log\frac{en}{s_i} }{|G \cap V|} \right) \leq e^{-t}
\]
Now note that $|G \cap V| \geq (\max_is_i) (n_0 - 1)(n_0 - 2) \cdots (n_0 - m + 1) \geq C n_0^{m-1} (\max_is_i)$. This gives us
\begin{align*}
	&\P\left( X_s \geq \| \mathbf{\bar{\Theta}}  \|_\infty + m!C\frac{t + \sum_{i=1}^m s_i \log\frac{en}{s_i} }{ n_0^{m-1} (\max_is_i)} \right) \leq e^{-t} \\
	\Rightarrow&\P\left( X_s \geq \| \mathbf{\bar{\Theta}}  \|_\infty + m!C\frac{t/n_0 + \sum_{i=1}^m \log\frac{en}{s_i} }{ n_0^{m-1} } \right) \leq e^{-t} \\
	\Rightarrow&\P\left( X_s \geq \| \mathbf{\bar{\Theta}}  \|_\infty + m!C\frac{t/n_0 + m\log\frac{n}{n_0} }{ n_0^{m-1} } \right) \leq e^{-t}.\\
\end{align*}
Further taking a union bound over $s_i \in \{ n_0, \ldots, n \}$ produces a $\frac{m}{n_0}\log n$ term, which is dominated by the $m \log\frac{n}{n_0}$, and we get
\[
	\P\left( \| \tilde{\mathbf{E}}_{\hat z} \|_\infty \leq \| \mathbf{\bar{\Theta}}  \|_\infty + m!C\frac{t + m\log\frac{n}{n_0} }{ n_0^{m-1} } \right) \geq 1-e^{-t}.
\]
Recall that
\[
	\langle \tilde{\mathbf{E}}_{\hat z}, \mathbf{E} \rangle \leq 2 \| \tilde{\mathbf{E}}_{\hat z} \|_F \sqrt{2m! \| \mathbf{\bar{\Theta}}  \|_\infty (t + n\log k + k^m)} + \frac{4m!}{3} \| \tilde{\mathbf{E}}_{\hat z} \|_\infty (t + n\log k + k^m)
\]
with probability at least $1-e^{-t}$. Combining this with our bound on $\| \tilde{\mathbf{E}}_{\hat z} \|_\infty$ and using AM-GM gives us


\[
	\langle \tilde{\mathbf{E}}_{\hat z}, \mathbf{E} \rangle \leq \frac{\|\tilde{\mathbf{E}}_{\hat z}\|_F^2}{16} + (m!)^2C \left( \| \mathbf{\bar{\Theta}}  \|_\infty (t + n\log k + k^m) + \frac{t + m\log\frac{n}{n_0}}{n_0^{m-1}} (t + n\log k + k^m)  \right)
\]
with probability at least $1 - 3e^{-t}$.

Now we proceed to obtain a bound in expectation from the above probability bound. In order to do so, note that by Lemma \ref{lem:quad}, we have
\begin{align*}
	\E\left[\langle \tilde{\mathbf{E}}_{\hat z}, \mathbf{E} \rangle - \frac{\|\tilde{\mathbf{E}}_{\hat z}\|_F^2}{16} \right] \leq& (m!)^2C\left( \| \mathbf{\bar{\Theta}}  \|_\infty(n\log k + k^m) + \frac{m\log\frac{n}{n_0}}{n_0^{m-1}}(n\log k + k^m)  \right) \\
	& + (m!)^2C\left( \| \mathbf{\bar{\Theta}}  \|_\infty + \frac{m\log\frac{n}{n_0}}{n_0^{m-1}} + \frac{n\log k + k^m}{n_0^{m-1}} \right) \\
	\leq& (m!)^2C\left( \| \mathbf{\bar{\Theta}}  \|_\infty(n\log k + k^m) + \frac{m\log\frac{n}{n_0}}{n_0^{m-1}}(n\log k + k^m)  \right)
\end{align*}

Recall that we have
\[
	\E\left[\langle \tilde{\mathbf{\Theta}}_{\hat z} - \mathbf{\bar{\Theta}} , \mathbf{E} \rangle - \frac{\| \tilde{\mathbf{\Theta}}_{\hat z} - \mathbf{\bar{\Theta}}  \|_F^2}{8}\right] \leq Cm!\| \mathbf{\bar{\Theta}}  \|_\infty n\log k
\]
and by definition, $\| \tilde{\mathbf{\Theta}}_{\hat z} - \mathbf{\bar{\Theta}}  \|_F \leq \| \hat{\mathbf{\Theta}}_{\hat z} - \mathbf{\bar{\Theta}}  \|_F$. Combining the above observation, we have
\begin{align*}
	\| \tilde{\mathbf{E}}_{\hat z} \|_F &= \| \hat{\mathbf{\Theta}}_{\hat z} - \tilde{\mathbf{\Theta}}_{\hat z} \|_F \\
					&\leq \| \hat{\mathbf{\Theta}}_{\hat z} - \mathbf{\bar{\Theta}}  \|_F + \| \tilde{\mathbf{\Theta}}_{\hat z} - \mathbf{\bar{\Theta}}  \|_F \\
					&\leq 2\| \hat{\mathbf{\Theta}}_{\hat z} - \mathbf{\bar{\Theta}}  \|_F.
\end{align*}
This gives us the required bound in expectation. 
\begin{align}\label{eq:control22}
	\E \left[ \langle \hat{\mathbf{\Theta}} - \mathbf{\bar{\Theta}} , \mathbf{E} \rangle  \right] \leq \frac38 \E \| \hat{\mathbf{\Theta}} - \mathbf{\bar{\Theta}}  \|_F^2 + (m!)^2C\left( \| \mathbf{\bar{\Theta}}  \|_\infty(n\log k + k^m) + \frac{m\log\frac{n}{n_0}}{n_0^{m-1}}(n\log k + k^m)  \right)
\end{align}

The statement of the proposition then follows immediately based on combing the above Equations~\ref{eq:control1} and \ref{eq:control22}.

\end{proof}

\begin{proof}[Proof of Lemma~\ref{lem:approx}]
For the sake of clarity, in this proof we will write $\mathbf{\Theta}$ and $f$ in place of $\mathbf{\bar{\Theta}} $ and $f_0$ respectively.  Recall that in the hypergraphon setting, $\mathbf{\Theta}$ is define by $\mathbf{\Theta}_j = \rho_n f({X}_j)$, and conditional on $\mathbf{\Theta}$, $\mathbf{\Theta}_{*}$ is deterministic. Instead of working directly with $\mathbf{\Theta}_{*}$, we define $\ddot{\mathbf{\Theta}}$ to be a blockwise average of $\mathbf{\Theta}$ on a balanced partition $z^*$, where the first $k-1$ blocks have $n_0$ elements, and the $k$th block contains $n - (k-1)n_0$. More precisely, define $z^*$ by
\[
	z^{-1}(a) = \{ i \in [n] : {X}_i = {X}_{(\ell)} \text{ for some } \ell \in [(a-1)n_0 + 1, an_0] \}
\]
when $a \in \{ 1, 2, \ldots, k-1 \}$, and for $a=k$ we have
\[
	z^{-1}(k) = \{ i \in [n] : {X}_i = {X}_{(\ell)} \text{ for some } \ell \in [(k-1)n_0 + 1, n] \}
\]

where ${X}_{(j)}$ denotes the $j$th order statistic of the vector ${X}$. Let $n = n_0k + r$, where $r \in \{ 0, 1, \ldots, n_0-1 \}$. Note that by construction, the first $k-1$ classes contain exactly $n_0$ elements, and the last one contains $n_0 + r$. For $a \in [k]^m$, we define $\eta_{a}^*$ to be the number of (unordered) hyperedges in the collection of classes $a$. Let $c_i$ be the number of times class $i$ appears in the vector $a$. Then
\[
	\eta_a^* = \binom{n_0 + r}{c_k} \prod_{i \in [k-1]} \binom{n_0}{c_i}
\]
Now we define the following blockwise average
\[
	\mathbf{Q}_a^* = \frac{1}{\eta_a^*} \sum_{j \in (z^*)^{-1}(a) \cap H} \mathbf{\Theta}_j = \frac{1}{\eta_a^*} \sum_{j \in (z^*)^{-1}(a) \cap H} \rho_n W({X}_j)
\]
Finally, we define $\ddot{\mathbf{\Theta}}$ to be given by $\ddot{\mathbf{\Theta}}_j = Q_{z(j)}^*$ for $j \in G$, and zero otherwise. Then we have,
\begin{align*}
\E\left[ \frac{1}{n^m} \| \mathbf{\Theta} - \ddot{\mathbf{\Theta}} \|_F^2 \right] &= \frac{1}{n^m} \sum_{a \in [k]^m} \E \sum_{j \in (z^*)^{-1}(a) \cap G} \left( \mathbf{\Theta}_j - \mathbf{Q}_a^* \right)^2 \\
			&= \frac{1}{n^m} \sum_{a \in [k]^m} \E \sum_{j \in (z^*)^{-1}(a) \cap G} \left( \rho_n f({X}_j) - \frac{1}{\eta_a^*} \sum_{u \in (z^*)^{-1}(a) \cap H} \rho_n f({X}_u) \right)^2 \\
			&= \frac{\rho_n^2}{n^m} \sum_{a \in [k]^m} \E \sum_{j \in (z^*)^{-1}(a) \cap G} \left( \frac{1}{\eta_a^*} \sum_{u \in (z^*)^{-1}(a)\cap H} (f({X}_j) - f({X}_u)) \right)^2 .
\end{align*}
Note that we can interpret the inner sum as an expectation and use Jensen's inequality to obtain
\[
	\left( \frac{1}{\eta_a^*} \sum_{u \in (z^*)^{-1}(a)\cap H} (f({X}_j) - f({X}_u)) \right)^2 \leq \frac{1}{\eta_a^*} \sum_{u \in (z^*)^{-1}(a)\cap H} (f({X}_j) - f({X}_u))^2.
\]
This then implies
\[
	\E\left[ \frac{1}{n^m} \| \mathbf{\Theta} - \ddot{\mathbf{\Theta}} \|_F^2 \right] \leq \frac{\rho_n^2}{n^m} \sum_{a \in [k]^m} \sum_{j \in (z^*)^{-1}(a) \cap G} \frac{1}{\eta_a^*} \sum_{u \in (z^*)^{-1}(a)\cap H} \E[(f({X}_j) - f({X}_u))^2]
\]

By leveraging the Lipschitz smoothness assumption in Equation~\ref{eq:smoothness}, the inequality $(\sum_i a_i)^2 \leq 2\sum_i a_i^2$, and Jensen's inequality respectively, we then have the following:
\begin{align*}
	\E[(f({X}_j) - f({X}_u))^2] &\leq M^2 \max_i \E\left[ ( |{X}_{j_1} - {X}_{u_1}| , \ldots , |{X}_{j_m} - {X}_{u_m}| )^2 \right] \\
						&\leq 2M^2\max_i   \left( \E \left[|{X}_{j_1} - {X}_{u_1}|^{2}\right],  \ldots,  \E\left[ |{X}_{j_m} - {X}_{u_m}|^{2}\right] \right) \\
						&\leq 2M^2 \max_i  \left( \E \left[|{X}_{j_1} - {X}_{u_1}|^2\right]+ \ldots + \E\left[ |{X}_{j_m} - {X}_{u_m}|^2\right]\right),
\end{align*}
Let $\sigma_x$ be such that ${X}_x = {X}_{(\sigma_x)}$. For example, $\sigma$ applied to the maximal element of ${X}$ would be 1. Since in the above expression, $j_i$ and $u_i$ belong to the same cluster for every $i$, by construction, we have $|\sigma_{j_i} - \sigma_{u_i}| \leq 2n_0$ for every $i$. By lemma~\ref{lem:klopp} and that fact that $k = \lfloor n/n_0 \rfloor$,
\[
	\E \left[|{X}_{j_i} - {X}_{u_i}|^2 \right] \leq \left( \frac{2n_0}{n} \right) ^{2} \leq 4\left( \frac{1}{k} \right) ^{2}
\]
which implies
\[
	\E[(f({X}_j) - f({X}_u))^2] \leq 8M^2 \left( \frac{1}{k} \right) ^{2}
\]

Putting it all together, we obtain
\[
	\E\left[ \frac{1}{n^m} \| \mathbf{\Theta} - \ddot{\mathbf{\Theta}} \|_F^2 \right] \leq 8 M^2 \rho_n^2 \left( \frac{1}{k^2} \right)
\]
which completes the proof.

\end{proof}

\section{Auxiliary Results}
In this section, we list auxiliary results that are used in the proofs of the main results. 

\begin{lemma}\label{lem:quad}
Let $X$ be a random variable such that its tail-decay satisfy: $\P(X > u^2 + Bu + C) \leq e^{-u}$ for all $u$. Then $\E[X] \leq C + 2 + B$.
\end{lemma}

\begin{proof}
By assumption $\P(X > u^2 + Bu + C) \leq e^{-u}$. We have
\[
	\E[X] \leq t^* + \int_{t^*}^\infty \P(X > t) dt
\]
We choose $t^* = C$ so that on $[t^*, \infty)$, $t = u^2 + Bu + C$ has at least one nonnegative solution and is necessarily increasing. Let $t = u^2 + Bu + C$.

\begin{align*}
	\E[X] &\leq C + \int_C^\infty \P(X > t) dt \\
		&= C + \int_C^\infty \P(X > u^2 + Bu + C) dt \\
		&\leq C + \int_0^\infty e^{-u} (2u + B) du \\
		&= C + 2 + B.
\end{align*}
\end{proof}
We also need the following result from~\cite{klopp2015oracle}. 
\begin{lemma}\label{lem:klopp}
Let $Z_1, \ldots, Z_n$ be i.i.d uniformly distributed random variables on $[0,1]$. Let $Z_{(i)}$ be the $i$-th order statistic of the above set of random variables. Then for any $n_0 \leq n$ and $0 \leq s < n_0$, we have
$$ \E \left(Z_{(i)} - Z_{(i+s)} \right)^2 = \frac{s(s+1)}{(n+1)(n+2)} \leq \left(\frac{n_0}{n}\right)^2.$$
\end{lemma}
The proof of the above lemma is straightforward and could be found in~\cite{klopp2015oracle}. Furthermore, since there are several different versions of Bernstein's inequality in the literature, we also state the version that we used in our proofs, below.
\begin{thm}[Bernstein's Inequality]\label{thm:bern}
Let $X_1, X_2, \ldots, X_N$ be zero-mean independent random variables such that $|X_i| \leq M$ almost surely. Then for all $t\geq 0$
\[
	\P\left( \sum_{i=1}^N X_i \geq \sqrt{2t \sum_{i=1}^N E[X_i^2] } +\frac{2M}{3}t \right) \leq e^{-t}.
\]
\end{thm}

\bibliographystyle{plainnat}
\bibliography{hypergraphon}

\begin{thebibliography}{77}
\providecommand{\natexlab}[1]{#1}
\providecommand{\url}[1]{\texttt{#1}}
\expandafter\ifx\csname urlstyle\endcsname\relax
  \providecommand{\doi}[1]{doi: #1}\else
  \providecommand{\doi}{doi: \begingroup \urlstyle{rm}\Url}\fi

\bibitem[Abbe(2017)]{abbe2017community}
Emmanuel Abbe.
\newblock Community detection and stochastic block models: recent developments.
\newblock \emph{The Journal of Machine Learning Research}, 18\penalty0
  (1):\penalty0 6446--6531, 2017.

\bibitem[Abbe and Sandon(2015)]{abbe2015community}
Emmanuel Abbe and Colin Sandon.
\newblock Community detection in general stochastic block models: Fundamental
  limits and efficient algorithms for recovery.
\newblock In \emph{Foundations of Computer Science (FOCS), 2015 IEEE 56th
  Annual Symposium on}, pages 670--688, 2015.

\bibitem[Abbe et~al.(2017)Abbe, Fan, Wang, and Zhong]{abbe2017entrywise}
Emmanuel Abbe, Jianqing Fan, Kaizheng Wang, and Yiqiao Zhong.
\newblock Entrywise eigenvector analysis of random matrices with low expected
  rank.
\newblock \emph{arXiv preprint arXiv:1709.09565}, 2017.

\bibitem[Agarwal et~al.(2005)Agarwal, Lim, Zelnik-Manor, Perona, Kriegman, and
  Belongie]{agarwal2005beyond}
S.~Agarwal, J.~Lim, L.~Zelnik-Manor, P.~Perona, D.~Kriegman, and S.~Belongie.
\newblock Beyond pairwise clustering.
\newblock \emph{IEEE Conference on Computer Vision and Pattern Recognition},
  2005.

\bibitem[Ahn et~al.(2018)Ahn, Lee, and Suh]{ahn2018hypergraph}
Kwangjun Ahn, Kangwook Lee, and Changho Suh.
\newblock Hypergraph spectral clustering in the weighted stochastic block
  model.
\newblock \emph{IEEE Journal of Selected Topics in Signal Processing},
  12\penalty0 (5):\penalty0 959--974, 2018.

\bibitem[Ahn et~al.(2019)Ahn, Lee, and Suh]{ahn2019community}
Kwangjun Ahn, Kangwook Lee, and Changho Suh.
\newblock Community recovery in hypergraphs.
\newblock \emph{IEEE Transactions on Information Theory}, 65\penalty0
  (10):\penalty0 6561--6579, 2019.

\bibitem[Airoldi et~al.(2013)Airoldi, Costa, and Chan]{airoldi2013stochastic}
Edo~M Airoldi, Thiago~B Costa, and Stanley~H Chan.
\newblock Stochastic blockmodel approximation of a graphon: Theory and
  consistent estimation.
\newblock In \emph{Advances in Neural Information Processing Systems}, pages
  692--700, 2013.

\bibitem[Aldous(1981)]{aldous1981representations}
David~J Aldous.
\newblock Representations for partially exchangeable arrays of random
  variables.
\newblock \emph{Journal of Multivariate Analysis}, 11\penalty0 (4):\penalty0
  581--598, 1981.

\bibitem[Angelini et~al.()Angelini, Caltagirone, Krzakala, and
  Zdeborov{\'a}]{angelini2015spectral}
Maria~Chiara Angelini, Francesco Caltagirone, Florent Krzakala, and Lenka
  Zdeborov{\'a}.
\newblock Spectral detection on sparse hypergraphs.
\newblock In \emph{2015 53rd Annual Allerton Conference on Communication,
  Control, and Computing (Allerton)}, pages 66--73. IEEE.

\bibitem[Battiston et~al.(2020)Battiston, Cencetti, Iacopini, Latora, Lucas,
  Patania, Young, and Petri]{battiston2020networks}
Federico Battiston, Giulia Cencetti, Iacopo Iacopini, Vito Latora, Maxime
  Lucas, Alice Patania, Jean-Gabriel Young, and Giovanni Petri.
\newblock Networks beyond pairwise interactions: {S}tructure and dynamics.
\newblock \emph{Physics Reports}, 2020.

\bibitem[Benson et~al.(2016)Benson, Gleich, and Leskovec]{benson2016higher}
Austin~R Benson, David~F Gleich, and Jure Leskovec.
\newblock Higher-order organization of complex networks.
\newblock \emph{Science}, 353\penalty0 (6295):\penalty0 163--166, 2016.

\bibitem[Bickel and Chen(2009)]{bickel2009nonparametric}
Peter~J Bickel and Aiyou Chen.
\newblock A nonparametric view of network models and {Newman--Girvan} and other
  modularities.
\newblock \emph{Proceedings of the National Academy of Sciences}, 106\penalty0
  (50):\penalty0 21068--21073, 2009.

\bibitem[Bonacich et~al.(2004)Bonacich, Holdren, and
  Johnston]{bonacich2004hyper}
P.~Bonacich, A.~C. Holdren, and M.~Johnston.
\newblock Hyper-edges and multidimensional centrality.
\newblock \emph{Social networks}, 26\penalty0 (3):\penalty0 189--203, 2004.

\bibitem[Borgs et~al.(2015{\natexlab{a}})Borgs, Chayes, Cohn, and
  Ganguly]{borgs2015consistent}
C.~Borgs, J.~T. Chayes, H.~Cohn, and S.~Ganguly.
\newblock Consistent nonparametric estimation for heavy-tailed sparse graphs.
\newblock \emph{arXiv preprint arXiv:1508.06675}, 2015{\natexlab{a}}.

\bibitem[Borgs et~al.(2015{\natexlab{b}})Borgs, Chayes, and
  Smith]{borgs2015private}
Christian Borgs, Jennifer Chayes, and Adam Smith.
\newblock Private graphon estimation for sparse graphs.
\newblock In \emph{Advances in Neural Information Processing Systems}, pages
  1369--1377, 2015{\natexlab{b}}.

\bibitem[Borgs et~al.(2019)Borgs, Chayes, Cohn, and Veitch]{borgs2019sampling}
Christian Borgs, Jennifer~T Chayes, Henry Cohn, and Victor Veitch.
\newblock Sampling perspectives on sparse exchangeable graphs.
\newblock \emph{The Annals of Probability}, 47\penalty0 (5):\penalty0
  2754--2800, 2019.

\bibitem[Campbell et~al.(2018)Campbell, Cai, and
  Broderick]{campbell2018exchangeable}
Trevor Campbell, Diana Cai, and Tamara Broderick.
\newblock Exchangeable trait allocations.
\newblock \emph{Electronic Journal of Statistics}, 12\penalty0 (2):\penalty0
  2290--2322, 2018.

\bibitem[Caron and Fox(2017)]{caron2017sparse}
Fran{\c{c}}ois Caron and Emily~B Fox.
\newblock Sparse graphs using exchangeable random measures.
\newblock \emph{Journal of the Royal Statistical Society: Series B (Statistical
  Methodology)}, 79\penalty0 (5):\penalty0 1295--1366, 2017.

\bibitem[Caron and Rousseau(2017)]{caron2017sparsity}
Fran{\c{c}}ois Caron and Judith Rousseau.
\newblock On sparsity and power-law properties of graphs based on exchangeable
  point processes.
\newblock \emph{arXiv preprint arXiv:1708.03120}, 2017.

\bibitem[Chan and Airoldi(2014)]{chan2014consistent}
Stanley Chan and Edoardo Airoldi.
\newblock A consistent histogram estimator for exchangeable graph models.
\newblock In \emph{International Conference on Machine Learning}, pages
  208--216, 2014.

\bibitem[Chatterjee(2015)]{chatterjee2015matrix}
Sourav Chatterjee.
\newblock Matrix estimation by universal singular value thresholding.
\newblock \emph{The Annals of Statistics}, 43\penalty0 (1):\penalty0 177--214,
  2015.

\bibitem[Chatterjee et~al.(2011)Chatterjee, Diaconis, and
  Sly]{chatterjee2011random}
Sourav Chatterjee, Persi Diaconis, and Allan Sly.
\newblock Random graphs with a given degree sequence.
\newblock \emph{Annals of Applied Probability}, 21\penalty0 (4):\penalty0
  1400--1435, 2011.

\bibitem[Chen et~al.(2019)Chen, Chi, Fan, and Ma]{Chen2019}
Yuxin Chen, Yuejie Chi, Jianqing Fan, and Cong Ma.
\newblock Gradient descent with random initialization: fast global convergence
  for nonconvex phase retrieval.
\newblock \emph{Mathematical Programming}, 176\penalty0 (1):\penalty0 5--37,
  Jul 2019.

\bibitem[Chien et~al.(2018)Chien, Lin, and Wang]{chien2018community}
I~Chien, Chung-Yi Lin, and I-Hsiang Wang.
\newblock Community detection in hypergraphs: {O}ptimal statistical limit and
  efficient algorithms.
\newblock In \emph{International Conference on Artificial Intelligence and
  Statistics}, pages 871--879, 2018.

\bibitem[Choi et~al.(2012)Choi, Wolfe, and Airoldi]{choi2012stochastic}
David~S Choi, Patrick~J Wolfe, and Edoardo~M Airoldi.
\newblock Stochastic blockmodels with a growing number of classes.
\newblock \emph{Biometrika}, 99\penalty0 (2):\penalty0 273--284, 2012.

\bibitem[Crane and Dempsey(2018)]{crane2018edge}
Harry Crane and Walter Dempsey.
\newblock Edge exchangeable models for interaction networks.
\newblock \emph{Journal of the American Statistical Association}, 113\penalty0
  (523):\penalty0 1311--1326, 2018.

\bibitem[Dempsey et~al.(2019)Dempsey, Oselio, and
  Hero]{dempsey2019hierarchical}
Walter Dempsey, Brandon Oselio, and Alfred Hero.
\newblock Hierarchical network models for structured exchangeable interaction
  processes.
\newblock \emph{arXiv preprint arXiv:1901.09982}, 2019.

\bibitem[Diao et~al.(2016)Diao, Guillot, Khare, and Rajaratnam]{diao2016model}
Peter Diao, Dominique Guillot, Apoorva Khare, and Bala Rajaratnam.
\newblock Model-free consistency of graph partitioning.
\newblock \emph{arXiv preprint arXiv:1608.03860}, 2016.

\bibitem[Duchenne et~al.(2011)Duchenne, Bach, Kweon, and
  Ponce]{duchenne2011tensor}
O.~Duchenne, F.~Bach, I.~Kweon, and J.~Ponce.
\newblock A tensor-based algorithm for high-order graph matching.
\newblock \emph{IEEE transactions on pattern analysis and machine
  intelligence}, 33\penalty0 (12):\penalty0 2383--2395, 2011.

\bibitem[Eldridge et~al.(2016)Eldridge, Belkin, and Wang]{eldridge2016graphons}
Justin Eldridge, Mikhail Belkin, and Yusu Wang.
\newblock Graphons, mergeons, and so on!
\newblock In \emph{Advances in Neural Information Processing Systems}, pages
  2307--2315, 2016.

\bibitem[Elek and Szegedy(2012)]{elek2012measure}
G{\'a}bor Elek and Bal{\'a}zs Szegedy.
\newblock A measure-theoretic approach to the theory of dense hypergraphs.
\newblock \emph{Advances in Mathematics}, 231\penalty0 (3-4):\penalty0
  1731--1772, 2012.

\bibitem[Florescu and Perkins(2016)]{florescu2016spectral}
L.~Florescu and W.~Perkins.
\newblock Spectral thresholds in the bipartite stochastic block model.
\newblock \emph{Conference on Learning Theory}, 2016.

\bibitem[Fortunato(2010)]{fortunato2010community}
Santo Fortunato.
\newblock Community detection in graphs.
\newblock \emph{Physics reports}, 486\penalty0 (3):\penalty0 75--174, 2010.

\bibitem[Gao et~al.(2015)Gao, Lu, and Zhou]{gao2015rate}
Chao Gao, Yu~Lu, and Harrison~H Zhou.
\newblock Rate-optimal graphon estimation.
\newblock \emph{The Annals of Statistics}, 43\penalty0 (6):\penalty0
  2624--2652, 2015.

\bibitem[Gao et~al.(2018)Gao, Ma, Zhang, and Zhou]{gao2018community}
Chao Gao, Zongming Ma, Anderson~Y Zhang, and Harrison~H Zhou.
\newblock Community detection in degree-corrected block models.
\newblock \emph{The Annals of Statistics}, 46\penalty0 (5):\penalty0
  2153--2185, 2018.

\bibitem[Ghoshal et~al.(2009)Ghoshal, Zlati{\'c}, Caldarelli, and
  Newman]{ghoshal2009random}
Gourab Ghoshal, Vinko Zlati{\'c}, Guido Caldarelli, and MEJ Newman.
\newblock Random hypergraphs and their applications.
\newblock \emph{Physical Review E}, 79\penalty0 (6):\penalty0 066--118, 2009.

\bibitem[Ghoshdastidar and
  Dukkipati(2017{\natexlab{a}})]{ghoshdastidar2015consistency}
Debarghya Ghoshdastidar and Ambedkar Dukkipati.
\newblock Consistency of spectral hypergraph partitioning under planted
  partition model.
\newblock \emph{The Annals of Statistics}, 45\penalty0 (1):\penalty0 289--315,
  2017{\natexlab{a}}.

\bibitem[Ghoshdastidar and
  Dukkipati(2017{\natexlab{b}})]{ghoshdastidar2016uniform}
Debarghya Ghoshdastidar and Ambedkar Dukkipati.
\newblock Uniform hypergraph partitioning: {P}rovable tensor methods and
  sampling techniques.
\newblock \emph{The Journal of Machine Learning Research}, 18\penalty0
  (1):\penalty0 1638--1678, 2017{\natexlab{b}}.

\bibitem[Gin{\'e} and Nickl(2015)]{gine2015mathematical}
Evarist Gin{\'e} and Richard Nickl.
\newblock \emph{Mathematical foundations of infinite-dimensional statistical
  models}, volume~40.
\newblock Cambridge University Press, 2015.

\bibitem[Goldenberg et~al.(2010)Goldenberg, Zheng, Fienberg, and
  Airoldi]{goldenberg2010survey}
Anna Goldenberg, Alice~X Zheng, Stephen~E Fienberg, and Edoardo~M Airoldi.
\newblock A survey of statistical network models.
\newblock \emph{Foundations and Trends in Machine Learning}, 2\penalty0
  (2):\penalty0 129--233, 2010.

\bibitem[Gowers(2007)]{gowers2007hypergraph}
W~Timothy Gowers.
\newblock Hypergraph regularity and the multidimensional szemer{\'e}di theorem.
\newblock \emph{Annals of Mathematics}, pages 897--946, 2007.

\bibitem[Harper and Konstan(2015)]{harper2015movielens}
F~Maxwell Harper and Joseph~A Konstan.
\newblock The movielens datasets: {H}istory and context.
\newblock \emph{{ACM} transactions on interactive intelligent systems
  {(TIIS)}}, 5\penalty0 (4):\penalty0 1--19, 2015.

\bibitem[Hoff et~al.(2002)Hoff, Raftery, and Handcock]{hoff2002latent}
Peter~D Hoff, Adrian~E Raftery, and Mark~S Handcock.
\newblock Latent space approaches to social network analysis.
\newblock \emph{Journal of the american Statistical association}, 97\penalty0
  (460):\penalty0 1090--1098, 2002.

\bibitem[Holland et~al.(1983)Holland, Laskey, and
  Leinhardt]{holland1983stochastic}
Paul~W Holland, Kathryn~Blackmond Laskey, and Samuel Leinhardt.
\newblock Stochastic blockmodels: {F}irst steps.
\newblock \emph{Social networks}, 1983.

\bibitem[Hoover(1979)]{hoover1979relations}
Douglas~N Hoover.
\newblock Relations on probability spaces and arrays of random variables.
\newblock \emph{Preprint, Institute for Advanced Study, Princeton, NJ}, 2,
  1979.

\bibitem[Ji and Jin(2016)]{ji2016}
Pengsheng Ji and Jiashun Jin.
\newblock Coauthorship and citation networks for statisticians.
\newblock \emph{The Annals of Applied Statistics}, 10\penalty0 (4):\penalty0
  1779--1812, 2016.

\bibitem[Kallenberg(1999)]{kallenberg1999multivariate}
Olav Kallenberg.
\newblock Multivariate sampling and the estimation problem for exchangeable
  arrays.
\newblock \emph{Journal of Theoretical Probability}, 1999.

\bibitem[Karwa and Petrovia(2016)]{karwa2016}
Vishesh Karwa and Sonja Petrovia.
\newblock Discussion of coauthorship and citation networks for statisticians.
\newblock \emph{The Annals of Applied Statistics}, 2016.

\bibitem[Ke et~al.(2019)Ke, Shi, and Xia]{ke2019community}
Zheng~Tracy Ke, Feng Shi, and Dong Xia.
\newblock Community detection for hypergraph networks via regularized tensor
  power iteration.
\newblock \emph{arXiv preprint arXiv:1909.06503}, 2019.

\bibitem[Kim et~al.(2018)Kim, Bandeira, and Goemans]{kim2018stochastic}
Chiheon Kim, Afonso~S Bandeira, and Michel~X Goemans.
\newblock Stochastic block model for hypergraphs: {S}tatistical limits and a
  semidefinite programming approach.
\newblock \emph{arXiv preprint arXiv:1807.02884}, 2018.

\bibitem[Kivel{\"a} et~al.(2014)Kivel{\"a}, Arenas, Barthelemy, Gleeson,
  Moreno, and Porter]{kivela2014multilayer}
Mikko Kivel{\"a}, Alex Arenas, Marc Barthelemy, James~P Gleeson, Yamir Moreno,
  and Mason~A Porter.
\newblock Multilayer networks.
\newblock \emph{Journal of complex networks}, 2\penalty0 (3):\penalty0
  203--271, 2014.

\bibitem[Klopp and Verzelen(2019)]{klopp2017optimal}
Olga Klopp and Nicolas Verzelen.
\newblock Optimal graphon estimation in cut distance.
\newblock \emph{Probability Theory and Related Fields}, pages 1--58, 2019.

\bibitem[Klopp et~al.(2017)Klopp, Tsybakov, and Verzelen]{klopp2015oracle}
Olga Klopp, Alexandre~B Tsybakov, and Nicolas Verzelen.
\newblock Oracle inequalities for network models and sparse graphon estimation.
\newblock \emph{The Annals of Statistics}, 45\penalty0 (1):\penalty0 316--354,
  2017.

\bibitem[Kolaczyk(2009)]{kolaczyk2009statistical}
Eric~D Kolaczyk.
\newblock {Statistical Analysis of Network Data: Methods and Models (Springer
  Series in Statistics)}.
\newblock 2009.

\bibitem[Krzakala et~al.(2013)Krzakala, Moore, Mossel, Neeman, Sly,
  Zdeborov{\'a}, and Zhang]{krzakala2013spectral}
Florent Krzakala, Cristopher Moore, Elchanan Mossel, Joe Neeman, Allan Sly,
  Lenka Zdeborov{\'a}, and Pan Zhang.
\newblock Spectral redemption in clustering sparse networks.
\newblock \emph{Proceedings of the National Academy of Sciences}, 110\penalty0
  (52):\penalty0 20935--20940, 2013.

\bibitem[Le et~al.(2017)Le, Levina, and Vershynin]{Lesparse2017}
Can~M. Le, Elizaveta Levina, and Roman Vershynin.
\newblock Concentration and regularization of random graphs.
\newblock \emph{Random Structures \& Algorithms}, 51\penalty0 (3):\penalty0
  538--561, 2017.

\bibitem[Lei and Rinaldo(2015)]{lei2015consistency}
Jing Lei and Alessandro Rinaldo.
\newblock Consistency of spectral clustering in stochastic block models.
\newblock \emph{The Annals of Statistics}, 43\penalty0 (1):\penalty0 215--237,
  2015.

\bibitem[Leurgans et~al.(1993)Leurgans, Ross, and
  Abel]{leurgans1993decomposition}
SE~Leurgans, RT~Ross, and RB~Abel.
\newblock A decomposition for three-way arrays.
\newblock \emph{SIAM Journal on Matrix Analysis and Applications}, 14\penalty0
  (4):\penalty0 1064--1083, 1993.

\bibitem[Lov{\'a}sz(2012)]{lovasz2012large}
L{\'a}szl{\'o} Lov{\'a}sz.
\newblock \emph{Large networks and graph limits}, volume~60.
\newblock American Mathematical Soc., 2012.

\bibitem[Lu and Zhou(2016)]{lu2016statistical}
Yu~Lu and Harrison~H Zhou.
\newblock {Statistical and Computational Guarantees of Lloyd's Algorithm and
  its Variants}.
\newblock \emph{arXiv preprint arXiv:1612.02099}, 2016.

\bibitem[Lunag{\'o}mez et~al.(2017)Lunag{\'o}mez, Mukherjee, Wolpert, and
  Airoldi]{lunagomez2017geometric}
Sim{\'o}n Lunag{\'o}mez, Sayan Mukherjee, Robert~L Wolpert, and Edoardo~M
  Airoldi.
\newblock Geometric representations of random hypergraphs.
\newblock \emph{Journal of the American Statistical Association}, 112\penalty0
  (517):\penalty0 363--383, 2017.

\bibitem[Michoel and Nachtergaele(2012)]{michoel2012alignment}
Tom Michoel and Bruno Nachtergaele.
\newblock Alignment and integration of complex networks by hypergraph-based
  spectral clustering.
\newblock \emph{Physical Review E}, 86\penalty0 (5):\penalty0 056111, 2012.

\bibitem[Ng and Murphy(2018)]{ng2018model}
Tin Lok~James Ng and Thomas~Brendan Murphy.
\newblock Model-based clustering for random hypergraphs.
\newblock \emph{arXiv preprint arXiv:1808.05185}, 2018.

\bibitem[Pal and Zhu(2019)]{pal2019community}
Soumik Pal and Yizhe Zhu.
\newblock Community detection in the sparse hypergraph stochastic block model.
\newblock \emph{arXiv preprint arXiv:1904.05981}, 2019.

\bibitem[Pensky(2019)]{pensky2019dynamic}
Marianna Pensky.
\newblock Dynamic network models and graphon estimation.
\newblock \emph{The Annals of Statistics}, 47\penalty0 (4):\penalty0
  2378--2403, 2019.

\bibitem[Rohe et~al.(2011)Rohe, Chatterjee, and Yu]{rohe2011spectral}
Karl Rohe, Sourav Chatterjee, and Bin Yu.
\newblock Spectral clustering and the high-dimensional stochastic blockmodel.
\newblock \emph{The Annals of Statistics}, pages 1878--1915, 2011.

\bibitem[Sharma et~al.(2020)Sharma, Patil, and Murty]{sharmac3mm}
Govind Sharma, Prasanna Patil, and M~Narasimha Murty.
\newblock {C3MM: Clique-Closure based Hyperlink Prediction}.
\newblock In \emph{Proceedings of the Twenty-Ninth International Joint
  Conference on Artificial Intelligence (IJCAI-20)}, 2020.

\bibitem[Stasi et~al.(2014)Stasi, Sadeghi, Rinaldo, Petrovic, and
  Fienberg]{stasi2014beta}
Despina Stasi, Kayvan Sadeghi, Alessandro Rinaldo, Sonja Petrovic, and Stephen
  Fienberg.
\newblock $\beta$ models for random hypergraphs with a given degree sequence.
\newblock In \emph{2014 21st International Conference on Computational
  Statistics}, page 593, 2014.

\bibitem[Turnbull et~al.(2019)Turnbull, Lunag{\'o}mez, Nemeth, and
  Airoldi]{turnbull2019latent}
Kathryn Turnbull, Sim{\'o}n Lunag{\'o}mez, Christopher Nemeth, and Edoardo
  Airoldi.
\newblock Latent space representations of {H}ypergraphs.
\newblock \emph{arXiv preprint arXiv:1909.00472}, 2019.

\bibitem[Veitch and Roy(2019)]{veitch2019sampling}
Victor Veitch and Daniel~M Roy.
\newblock Sampling and estimation for (sparse) exchangeable graphs.
\newblock \emph{The Annals of Statistics}, 47\penalty0 (6):\penalty0
  3274--3299, 2019.

\bibitem[Wolfe and Olhede(2013)]{wolfe2013nonparametric}
Patrick~J Wolfe and Sofia~C Olhede.
\newblock Nonparametric graphon estimation.
\newblock \emph{arXiv preprint arXiv:1309.5936}, 2013.

\bibitem[Yang et~al.(2014)Yang, Han, and Airoldi]{yang2014nonparametric}
Justin Yang, Christina Han, and Edoardo Airoldi.
\newblock Nonparametric estimation and testing of exchangeable graph models.
\newblock In \emph{Artificial Intelligence and Statistics}, pages 1060--1067,
  2014.

\bibitem[Zhang et~al.(2018)Zhang, Cui, Jiang, and Chen]{zhang2018beyond}
Muhan Zhang, Zhicheng Cui, Shali Jiang, and Yixin Chen.
\newblock {Beyond link prediction: Predicting hyperlinks in adjacency space}.
\newblock In \emph{Proceedings of the AAAI Conference on Artificial
  Intelligence}, volume~32, 2018.

\bibitem[Zhang et~al.(2017)Zhang, Levina, and Zhu]{zhang2017estimating}
Yuan Zhang, Elizaveta Levina, and Ji~Zhu.
\newblock Estimating network edge probabilities by neighbourhood smoothing.
\newblock \emph{Biometrika}, 104\penalty0 (4):\penalty0 771--783, 2017.

\bibitem[Zhao(2015)]{zhao2015hypergraph}
Y.~Zhao.
\newblock Hypergraph limits: {A} regularity approach.
\newblock \emph{Random Structures \& Algorithms}, 47\penalty0 (2):\penalty0
  205--226, 2015.

\bibitem[Zheng et~al.(2010)Zheng, Cao, Zheng, Xie, and
  Yang]{zheng2010collaborative}
Vincent Zheng, Bin Cao, Yu~Zheng, Xing Xie, and Qiang Yang.
\newblock Collaborative filtering meets mobile recommendation: {A}
  user-centered approach.
\newblock In \emph{Proceedings of the AAAI Conference on Artificial
  Intelligence}, volume~24, 2010.

\bibitem[Zlati{\'c} et~al.(2009)Zlati{\'c}, Ghoshal, and
  Caldarelli]{zlatic2009hypergraph}
Vinko Zlati{\'c}, Gourab Ghoshal, and Guido Caldarelli.
\newblock Hypergraph topological quantities for tagged social networks.
\newblock \emph{Physical Review E}, 80\penalty0 (3):\penalty0 036118, 2009.

\end{thebibliography}
\end{document}